\documentclass[nohyperref]{article}

\usepackage{arxiv}

\usepackage{microtype}
\usepackage{graphicx}
\usepackage{subfigure}
\usepackage{wrapfig}
\usepackage{booktabs}
\usepackage{hyperref}

\usepackage{amsmath}
\usepackage{amssymb}
\usepackage{mathtools}
\usepackage{amsthm}

\newtheorem{definition}{Definition}[section]

\def\eqref#1{(\ref{#1})}
\def\R{\mathbb R}

\def\1{\mathbbm{1}}

\usepackage[textsize=tiny]{todonotes}
\usepackage[capitalize,noabbrev]{cleveref}

\title{Understanding out-of-distribution accuracies through quantifying difficulty of test samples}

\author{
 Berfin \c{S}im\c{s}ek\thanks{Most of the work was performed while Berfin \c{S}im\c{s}ek was interning at Meta AI.} \\
  École Polytechnique Fédérale de Lausanne\\
  \texttt{berfin.simsek@epfl.ch} \\
   \And
 Melissa Hall \\
  Meta AI Research \\
  \texttt{melissahall@fb.com} \\
  \And
 Levent Sagun \\
  Meta AI Research \\
  \texttt{leventsagun@fb.com} \\
}

\begin{document}
\maketitle

\begin{abstract}
Existing works show that although modern neural networks achieve remarkable generalization performance on the in-distribution (ID) dataset, the accuracy drops significantly on the out-of-distribution (OOD) datasets \cite{recht2018cifar, recht2019imagenet}.
To understand why a variety of models consistently make more mistakes in the OOD datasets, we propose a new metric to quantify the difficulty of the test images (either ID or OOD) that depends on the interaction of the training dataset and the model.
In particular, we introduce \textit{confusion score} as a label-free measure of image difficulty which quantifies the amount of disagreement on a given test image based on the class conditional probabilities estimated by an ensemble of trained models.

Using the confusion score, we investigate CIFAR-10 and its OOD derivatives.
Next, by partitioning test and OOD datasets via their confusion scores, we predict the relationship between ID and OOD accuracies for various architectures. This allows us to obtain an estimator of the OOD accuracy of a given model only using ID test labels.
Our observations indicate that the biggest contribution to the accuracy drop comes from images with high confusion scores.
Upon further inspection, we report on the nature of the misclassified images grouped by their confusion scores: \textit{(i)} images with high confusion scores contain \textit{weak spurious correlations} that appear in multiple classes in the training data and lack clear \textit{class-specific features}, and \textit{(ii)} images with low confusion scores exhibit spurious correlations that belong to another class, namely \textit{class-specific spurious correlations}.
\end{abstract}

\section{Introduction}
\label{sec:Intro}

Recent progress in out-of-distribution (OOD) detection (and generation) has sped up developments in the quantified understanding of OOD~\cite{salehi2021unified}.
In particular, utilization of ensembles has shown improved accuracy in test sets for image classification problems \cite{lakshminarayanan2016simple, fort2021exploring}. The critical role of ensembling in accuracy improvement in test datasets can also be seen through the way it reduces NTK fluctuations due to initialization which contributes to accuracy gains for ID datasets as well~\cite{jacot2018neural, geiger2020scaling}. In this work, we utilize ensembles to take a deeper look at the structure of both ID and OOD datasets and further attempt to identify the sources of failures for image classification tasks via neural networks. With the use of entropy of the class conditional probabilities of test images estimated by ensembles of networks, we assign a \textit{confusion score} to each sample without the need of any labels.
We find that the majority of the accuracy drop in the OOD datasets in comparison to the ID dataset comes from samples with high confusion scores. In these images, the networks tend to vote between several classes relying less on the extracted features. In contrast, mistakes on images with low confusion scores are due to the fact that networks in the ensemble consistently predict a class other than the label. This may be due to label corruption or class-specific spurious correlations.

The confusion score quantifies how much agreement there is on a single image given the features extracted from the \textit{training dataset} by the networks in the ensemble.
Networks extract features between input images and class labels that are predictive of the class labels in the training data. This feature extraction procedure implicitly encodes a certain input-output correlation learning which also includes \textit{spurious correlations}.
The link between ensembling and feature learning with input-output correlations due to the compositional structure of standard datasets used for image recognition benchmarks is formalized in~\cite{allen2020towards}. In our work, we focus on how ensembles inform us about datasets through removing the noise and chance factor that may be introduced by the initialization that is independent of feature learning on the training set.

In test images, class specific features may not be pronounced due to the lack of similar samples in the training data. In such cases, we speculate that networks in the ensemble may extrapolate in an uninformed way towards regions of the input space that leverage correlations such as color or texture that are learned from the training data instead of using class specific features.

Inductive biases fall short in coming to the rescue, since the fundamental problem is that the objective of networks is to map the input images to the class labels in the classification tasks. We discuss these observations in more detail in Section~\ref{sec:interpretations}.

\begin{figure}[t!]%{r}{0.45\textwidth}
    \begin{center}
        \includegraphics[width=0.55\textwidth]{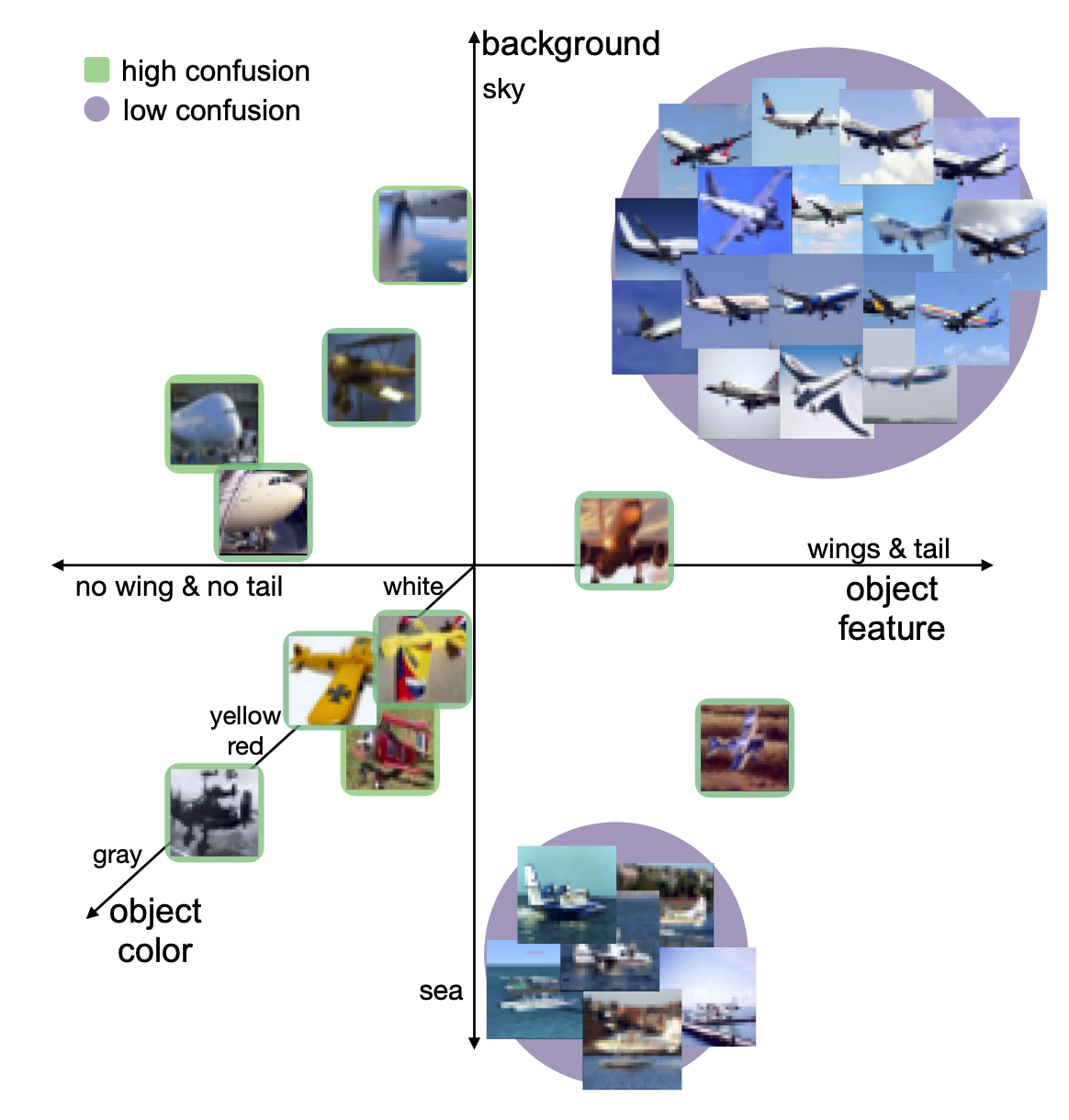}
      \end{center}
      \vspace{-0.3cm}
      \caption{\textbf{A schematic categorization of CIFAR-10 test images for the plane class with respect to the modalities extracted from the training data. Purple clusters: low confusion score, green frame: high confusion score.}
      Purple clusters represent images with lowest confusion scores: (i) majority of plane images exhibit features for planes clearly (open wings and wide tail), they also come with blue background; (ii) planes on the sea represent samples with low confusion score that are spuriously correlated with the ship class. These examples are not dominant in the dataset and form a small cluster.
      Green images are picked among the samples with highest confusion scores. They are difficult to classify either for not having clear object features (upper left quarter) or due to an atypical object or background color distribution (bottom half).
      \label{fig:plane-schem}
      }
\end{figure}

\noindent \textbf{Background and motivation:}
Much of the recent improvements in the state of the art accuracies in deep learning challenges come from scaling up the network size \cite{bahri2021explaining}. For vision datasets, along with the traditional convolutional networks \cite{kolesnikov2020big}, vision transformers and MLP-mixers achieve best accuracies with small differences despite having fundamentally different inductive biases \cite{zhai2021scaling, tolstikhin2021mlp}. Despite substantial differences in the architecture and the optimization methods, these models make systematic mistakes \cite{geirhos2021partial} and their accuracies decrease in a linear way on OOD datasets \cite{recht2018cifar, recht2019imagenet}. Moreover, evaluating simpler neural networks and classic machine learning models like random forests and k-NN's on both ID and OOD datasets show a linear trend of accuracies, suggesting that the systematic prediction trend carries over to a huge variety of models \cite{miller2021accuracy}. Also, vision transformers and convolutional networks make similar errors \cite{tuli2021convolutional, madan2021small} which indicate that they may be learning similar features and may be susceptible for memorizing similar correlations from the training dataset. These findings motivate us to study the structure of data from the point of view-of-the model.

\noindent \textbf{Contributions:} We present a label-free method to measure how confusing a test image is (either ID or OOD) given the modalities (i.e., features and correlations) leveraged from the training data. Our contributions can be summarized as:

\begin{itemize}
    \item We introduce a model based confusion score for any (ID or OOD) test sample.
    \item We find that OOD datasets have larger mean values of confusion score than the ID test dataset. Binning the confusion scores, we observe that the lowest confusion group in the ID dataset has a significantly higher participation ratio in the overall dataset compared to the OOD datasets. Because the error rates in higher confusion groups are increasingly higher, we identify that a distribution shift towards the high confusion groups in the OOD datasets is a major source of accuracy drop.
    \item We further inspect the nature of high and low confusion score samples which enables introducing two forms of spurious correlations: class-specific spurious correlations and weak spurious correlations.
    \item Using the confusion scores across an OOD dataset, we get a partition which enables us to predict OOD accuracies across a range of models without using the labels of the OOD datasets (but with using the labels in the test dataset).
\end{itemize}

The key motivator for the confusion score is its reliance on class-conditional probabilities that allow us to measure the extent to which correlations extracted from the training data are present in a single image.

Note also that we do not use any label information to calculate the confusion scores so our method can be employed for large unlabeled test datasets in the wild.

Our hypothesis: A neural network learns certain object features and correlations from the training data where some correlations may be spurious. Spurious correlations mainly take two forms\footnote{Spurious correlations may indeed come in forms that are not indicated here. Missing image parts, shifts in perspectives, labeling issues, even lack of documentation of how labeling is carried out can contribute to measurements of spurious correlations. In this work, we assume a data sampling process with consistent labeling practices, yet question the assumption that the full datasets are sampled iid. CIFAR-10 and its associated OOD datasets fit into this framework but arbitrary datasets may not.}: (1) \textit{class-specific spurious correlations:} image statistics that are correlated with exactly one other class than the object class (ex. sea background with the ship class),
(2) \textit{weak spurious correlations:} image statistics that are correlated with several classes (ex. white background with the car, ship, and deer class). Ensembling precisely helps on the population exhibiting weak spurious correlations!
Now the question is: how can we find the strong and weak forms of spurious correlations? Is it actually possible to find them?

\begin{figure*}[t!]
   \begin{center}
   \includegraphics[width=0.85\textwidth]{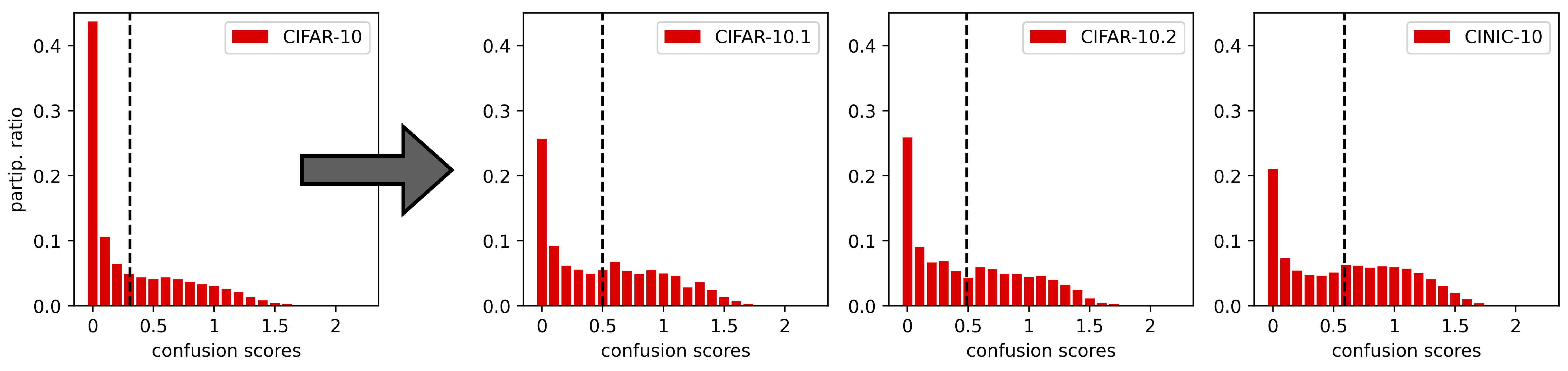} \\
   \includegraphics[width=0.42\textwidth]{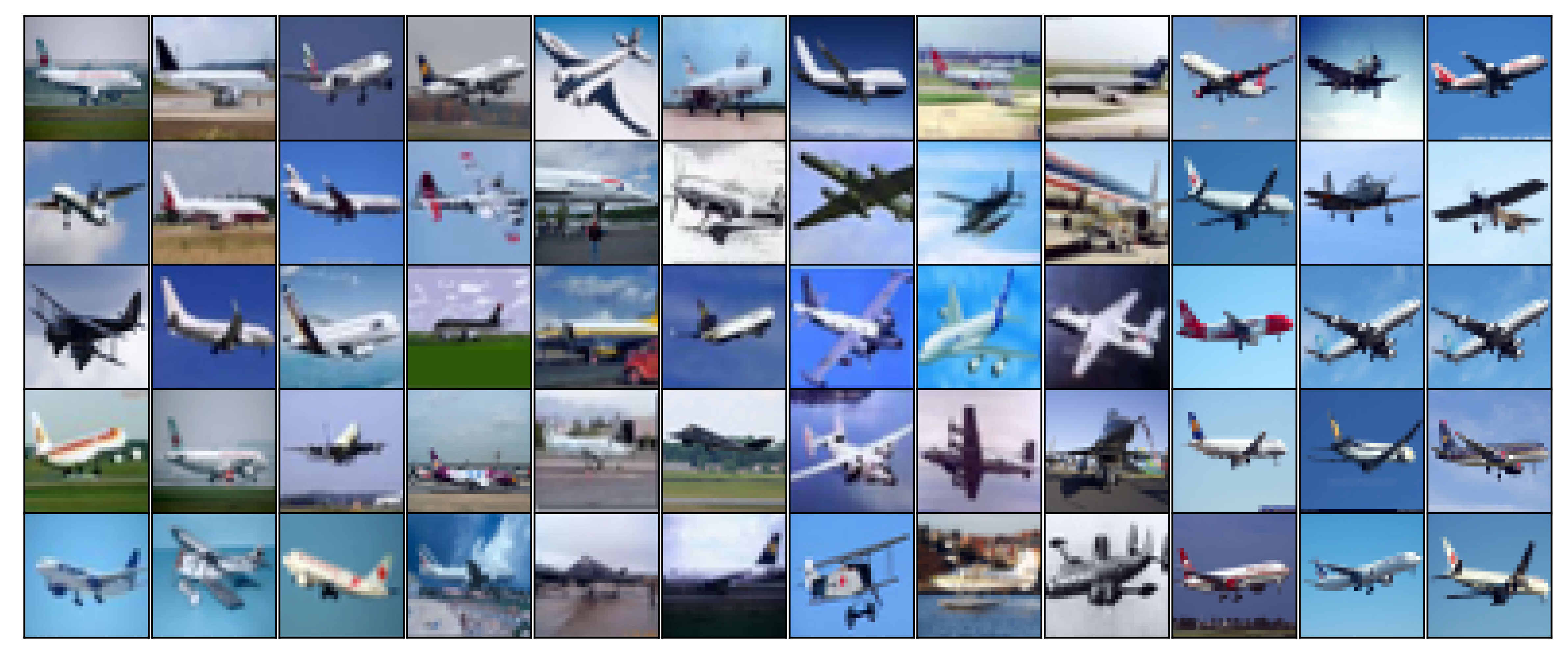} \hspace{3mm}
   \includegraphics[width=0.42\textwidth]{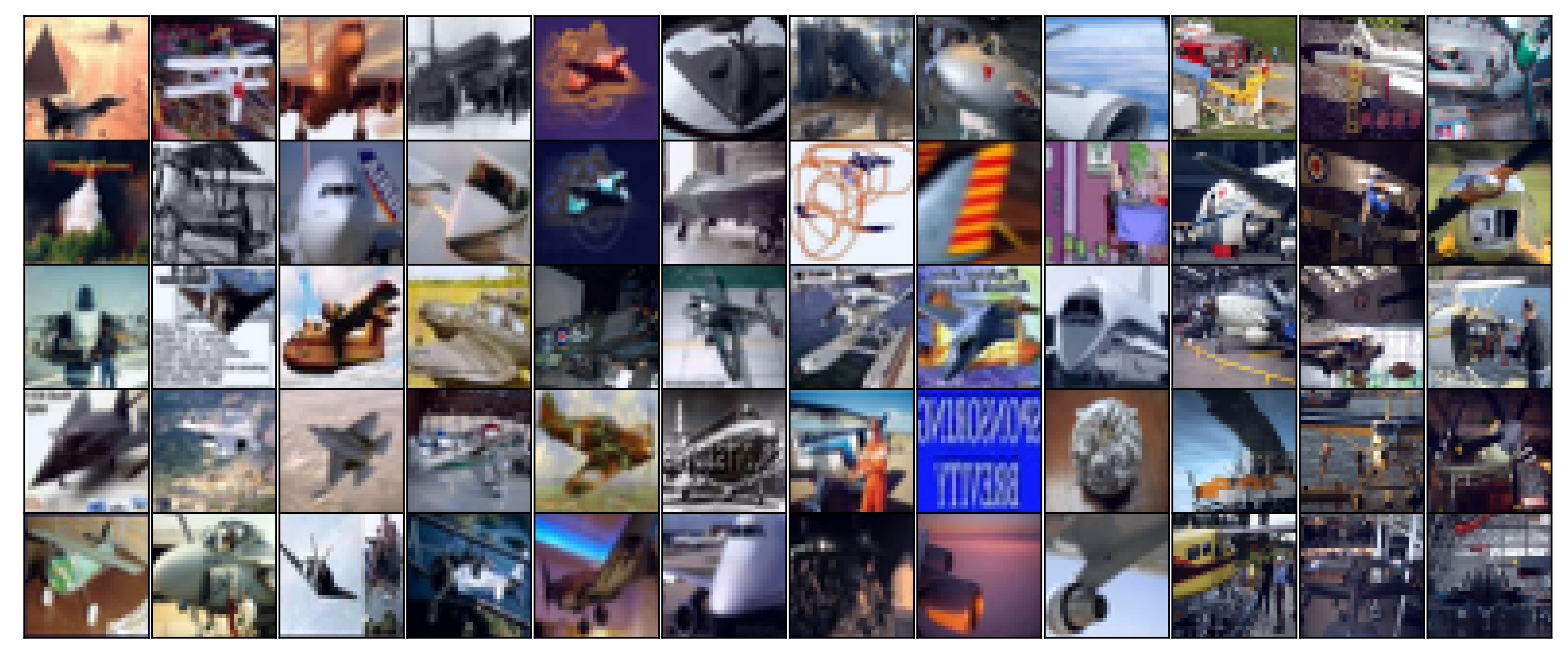} \\
   {\scriptsize \textbf{(a)} lowest confusion score images \hspace{40mm} \textbf{(b)} highest confusion score images}
   \end{center}
  \vspace{-0.3cm}
  \caption{\label{fig:confusion-dists} \textbf{Top: Distribution of the confusion scores.} Mean confusion scores are
  $0.357, 0.550, 0.538, 0.640$ for the CIFAR-10 test, CIFAR-10.1, CIFAR-10.2, and CINIC-10 datasets respectively. In the OOD datasets, the participation ratio of the images with low confusion score decreases, and the mean confusion score increases. The most confusing dataset above is the CINIC-10 since it has the highest mean confusion score. \textbf{Bottom: Extremely low and high confusion score images the from CIFAR-10 test dataset (ID) and three OOD datasets (CIFAR-10.1 \cite{recht2019imagenet}, CIFAR-10.2 \cite{lu2020harder}, and CINIC-10 \cite{darlow2018cinic}); three columns each.} Low confusion score images usually exhibit typical background, typical texture, and standard object features, but also mislabeled images and those with class-specific spurious correlations (see Figure~\ref{fig:sp-corr} for the examples of the latter groups). High confusion score images include instances of gray scale images, unusual background or object color, and objects that are missing class specific features. (e.g. plane images without neither wings nor tail, see Figure~\ref{fig:plane-schem}).}
\end{figure*}

\subsection{Related Work}

The authors of \cite{recht2018cifar} indicate that grouping the datasets into two categories as easy and hard may help explain the collinear relationship between ID and OOD accuracies. Building up on this observation, the authors of \cite{mania2020classifier} propose a candidate theory of the collinearity based on the assumption that the simple networks are rarely correct on the images which are misclassified by the complex networks (that the dominance probabilities are low in their terminology). However, the authors of \cite{sagawa2020investigation} demonstrate that overparameterization exacerbates performance on minority groups due to memorization, i.e. underparameterized networks have higher accuracies than overparameterized ones in this subpopulation (i.e. high dominance probability). On the other hand, the authors of \cite{andreassen2021evolution} find that pretrained models break the collinearity during fine-tuning since these models benefit from enormous image diversity in the training data. The authors of \cite{baldock2021deep} propose `prediction depth' for measuring image difficulty, but it is not clear how this measure can be used to compare the architectures with a large range of depths.

Entropy scores coming from the predictions of a bag of networks have previously been used for uncertainty predictions for out-of-distribution detection where the task is to predict whether the test image belong to a new class that has never been seen during training \cite{lakshminarayanan2016simple, belghazi2021classifiers}.
Another line of work focuses on data pruning where the goal is to reduce the number of training data points without incurring a drop in test accuracy \cite{toneva2018empirical, paul2021deep}. These works show that CIFAR-10 training data can be pruned up to $\%40$ either by using the forgetting scores (number of flips from correct to mistake during training) or the input gradients of several networks during early dynamics of training. The authors of \cite{agarwal2020estimating} identifies that images with low variance of gradients are typical whereas images with high variance of gradients exhibit atypical features.

\begin{figure*}[t!]
  \centering
  \includegraphics[width=0.65\textwidth]{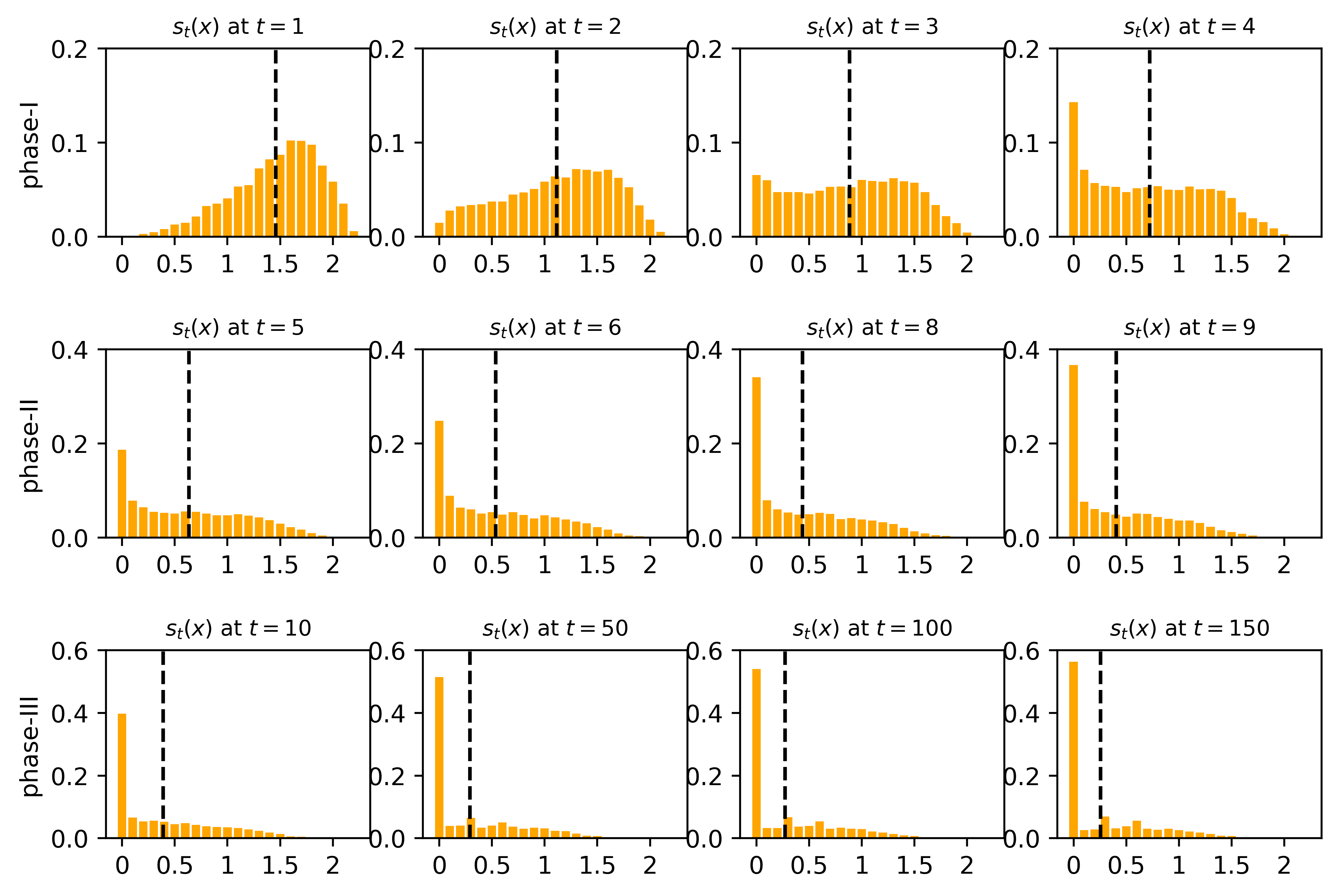} \hspace{0.5cm}
  \includegraphics[width=0.3\textwidth]{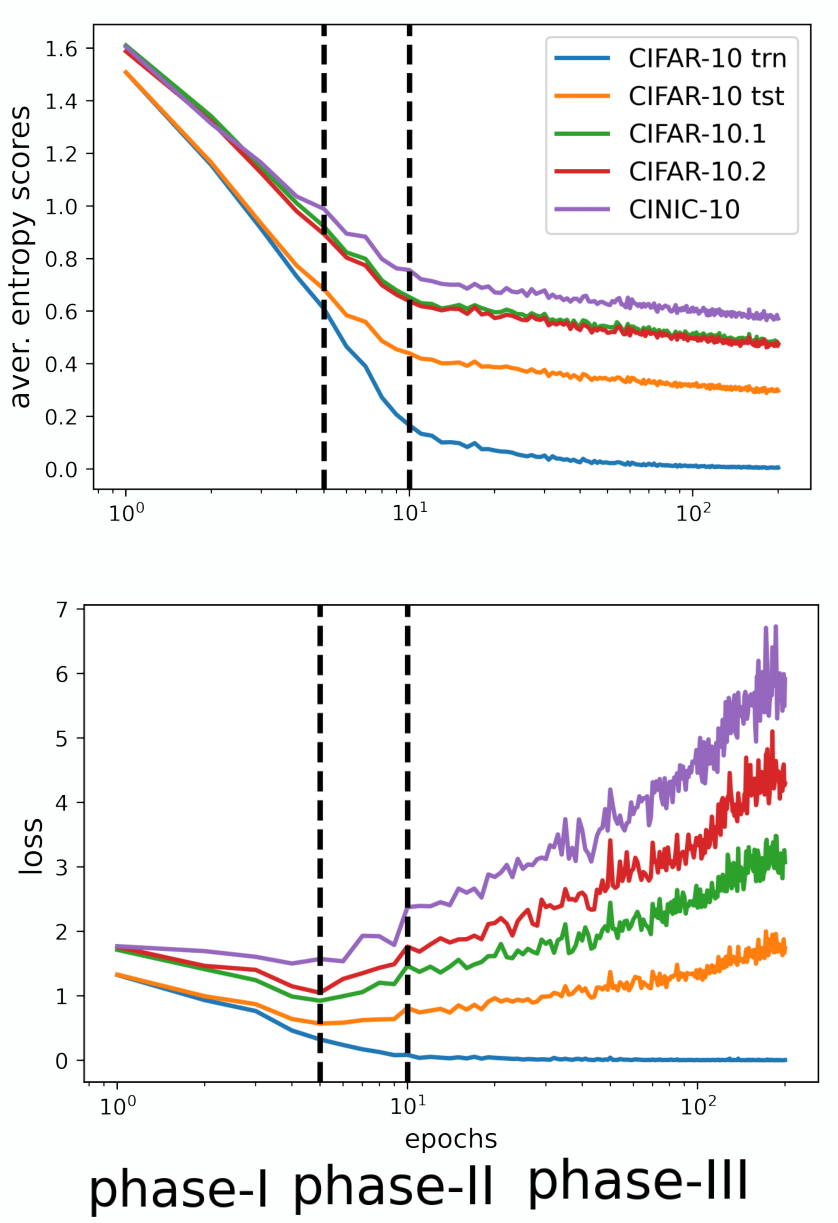}
  \centering
  \vspace{-0.3cm}
  \caption{\label{fig:time-eval-entrop-scores} \textbf{Left: Evolution of the entropy scores $s_t(x)$ through \textit{three} phases of training on the CIFAR-10 test dataset; \textit{phase-I:} fast early learning, \textit{phase-II:} slow early learning, \textit{phase-III:} memorization.} \textit{phase-I:} After the first epoch, we see that the majority of the samples are guessed randomly between $10$ classes, therefore reaching entropy levels close to $\log(10) \approx 2.3$. In the following few epochs, the easiest samples are assigned high confidence, therefore the bulk of the spectrum shifts to left rapidly. \textit{phase-II:} Starting from epoch $5$, the medium/high entropy samples possibly cycle among the entropy groups, yet the major qualitative change is the rapid increase of the population of the lowest entropy group. \textit{phase-III:} Starting from epoch $10$, the confidences increase due to the increasing weights trying to minimize the softmax objective. \textbf{Right: The average loss and average entropy scores during training.} We can identify the markers of three phases in the loss figure: in the first phase both the training and test losses decrease, in the second phase the training loss decreases whereas the test losses increase, and in the third phase, we see the memorization phase as the network interpolates the training data at the cost of assigning high confidence predictions to the regions of the input space.
  }
\end{figure*}

\section{Foundations \& Methodology}
\label{sec:foundations}

\begin{table}[t!]
\caption{Classification accuracies at epoch $t$ (averaged over $10$ seeds, see Appendix Fig.~\ref{fig-app:scores-during-training} for all $t$) \label{table:test-accs}}
\label{sample-table}
% \vskip 0.1in
\begin{center}
\begin{small}
\begin{sc}
\begin{tabular}{lcccr}
\toprule
Dataset & $t=5$ & $t=10$ & $t=50$ & $t=100$ \\
\midrule
CIFAR-10 trn & 0.91 & 0.98 & 1.00 & 1.00  \\
CIFAR-10 tst & 0.81 & 0.82 & 0.83 & 0.83 \\
CIFAR-10.1 & 0.68 & 0.70 & 0.70 & 0.71 \\
CIFAR-10.2 & 0.66 & 0.67 & 0.68 & 0.69 \\
CINIC-10 & 0.56 & 0.57 & 0.57 & 0.57 \\
\bottomrule
\end{tabular}
\end{sc}
\end{small}
\end{center}
% \vskip -0.1in
\end{table}

We aim to quantify the difficulty of a test image given the images in the training dataset.
In the best case scenario, a supervised learning algorithm learns the correlation between the input images and their class labels perfectly and cannot possibly learn whether the learned correlation is actually semantically related to the class label or not.
In the standard machine learning datasets such as CIFAR-10, the object (or similar) semantically associated with the class label is subject to deformations such as rotation, occlusion, color, or background change (see Figure~\ref{fig:plane-schem}). Therefore, when the object related features are deformed, the network is forced to leverage other modalities present in the samples of a class that are not semantically related the class label or the object.

\begin{definition}\label{def:class-specific-sp} The modalities\footnote{In this paper, modalities refer to two forms of image properties: (i) features such as wings or tail in the plane class; (ii) correlations such as object colors, background colors, or texture. By writing `the modalities that are not semantically related to any of the classes' we refer to the correlations, excluding class-specific features.} in the instances of a specific class that are not semantically related to the class and that correlate with the class label significantly more than the other class labels are referred to as \textbf{\emph{class-specific spurious correlations}}.
\end{definition}

The camel-cow binary classification problem presents a typical example of the class-specific spurious correlations \cite{arjovsky2019invariant}. %\levent{explain the problem}.
In the training dataset, camel images almost always exhibit sandy background, whereas cow images almost always exhibit green background. Extracting only the background color in this extreme case is sufficient for high accuracy on the training dataset. In test time, however, the networks then misclassify an image of a cow on a beige background. `Sandy/Beige background' is one type of a class-specific spurious correlation associated with the camel class, since this modality correlates with the camel class significantly more than the cow class.
This form of spurious correlation is well-studied in the binary classification setup \cite{sagawa2020investigation, arjovsky2019invariant}. In the multiclass classification setup, we additionally identify a new form of spurious correlations taking place:

\begin{definition}\label{def:weak-sp} The modalities in the instances that are not semantically related to any of the classes but that correlate with several class labels are referred to as \textbf{\emph{weak spurious correlations}}.
\end{definition}

As an example, many classes have a weak spurious correlation with a white background; a non-negligible proportion of the car, ship, bird, and deer class instances exhibit a clear white background (see Fig.~\ref{fig:low-high-entropy-app1} in the Appendix).

While weak spurious correlations and class-specific spurious correlations are observed in images from all ranges of confusion scores, we notice that weak-spurious correlations tend to occur more among samples with high confusion scores and vice versa for class-specific correlations, as exemplified in in Figure~\ref{fig:confusion-dists}. We also note that the definitions of two types of spurious correlations (Def.~\ref{def:class-specific-sp} and ~\ref{def:weak-sp}) are intentionally dataset-agnostic to allow for broader application.

\noindent \textbf{Confusion score:} To measure the presence of class-related correlations, we define \textit{confusion scores} based on the class conditional entropies of a bag of ResNets during training. We calculate a \textit{confusion score} for each test image in ID and OOD datasets. In particular, let $ f_t^{(i)} : \R^D \to \R^C $ be a network function that maps images ($D$-dimensional) to class conditional probabilities ($ C $ classes) at epoch $t$. For a given test image $ x \in \R^D $ we calculate the entropy scores at time $t$ as follows
\begin{align*}
    p_t(x) &= \frac{1}{M} \sum_{i=1}^M f_t^{(i)}(x) \\
    s_t(x) &= \sum_{j=1}^C -\log(p_t^j(x)) p_t^j(x)
\end{align*}
where $ M $ is the number of networks and $p_t(x)$ is the average class conditional probabilities of the ensemble. However, we observed that the entropy score of a single sample $s_t(x)$ fluctuates wildly during training especially for high confusion score samples (see Appendix Fig.~\ref{fig-app:entropy-instable-training}). This is likely due to (i) the differences in the speed of learning of the ensemble on high confusion score samples, and (ii) flipping the decision on the samples from correct to incorrect mostly in the phase-III of training as observed in a single network via the so-called `forgetting' scores in \cite{toneva2018empirical}. To remove the noise coming from a single epoch, we average the entropy scores over training
\begin{align*}
    s(x) = \frac{1}{T} \sum_{t=1}^T s_t(x)
\end{align*}
which we call the \textit{confusion score} of a test image $x$, where $T$ is the number of epochs.

\subsection{Three phases of learning \& evolution of the entropy scores}

Neural networks trained with a variant of gradient descent learn modalities of increasing complexity during training which is a phenomenon known as simplicity bias \cite{kalimeris2019sgd, rahaman2019spectral}. Similarly, from the dataset point of view, the easy images in the training dataset are the samples that are learned the fastest \cite{pezeshki2020gradient}. The test images that share similar simple features/correlations that are learned early in training by the network are therefore assigned low entropy scores in early training. Following \cite{baity2018comparing}, we identify the end of the early learning phase as the epoch where both the training and test accuracies improve only marginally afterward (see Table~\ref{table:test-accs}, Fig.~\ref{fig-app:scores-during-training}) which corresponds to the epoch where the both the average entropy score and the loss curves change slope (Fig.~\ref{fig:time-eval-entrop-scores}).

In the later phase of training, the test accuracies increase only a marginally (see Table~\ref{table:test-accs}). We call this phase \textit{memorization}, since $98\%$ of the training dataset samples are already classified accurately at epoch $t=10$. Furthermore, the network is forced to memorize difficult training samples either due to the lack of similar samples in the training dataset or some type of data corruption. Note that this phase is still beneficial for the test accuracies for both ID and OOD datasets therefore we view it as a benign memorization mechanism~\cite{feldman2020does}.
%Unlike in \citet{baity2018comparing},
We further split early learning into two phases based on the change of speed in learning in the first few epochs compare to the next few epochs as measured by the loss and accuracy metrics. The separating epoch is defined as where the test losses start to increase despite that the training loss keeps decreasing. We call the phase-I fast early learning, as training in this phase improves the average loss rapidly in both train and test datasets. We call the phase-II slow early learning, as the average loss increases for the test datasets due to overconfidence kicking in but the training accuracy still improves considerably from $91\%$ to $98\%$ (see Table~\ref{table:test-accs}).
The change in learning behavior as observed through the average metrics for the training and test datasets in these two phases is more pronounced from the view of individual data points. We observe a significant change in the movement of the entropy score distributions (Figure~\ref{fig:time-eval-entrop-scores}): (i) phase-I, a large proportion of high entropy samples quickly achieve lower entropy scores (not yet the lowest possible though), (ii) phase-II, further corrections of the entropy scores for the high entropy samples take place, resulting in a more stable subpopulation participation ratio for the lowest entropy group.

By averaging the entropy scores during training, we capture the information on learning order. For example, the first samples achieving low entropy scores exhibit simple modalities due to simplicity bias. Therefore those are the least confusing ones for the models.

\begin{figure}[t!]
   \begin{center}
   \includegraphics[width=0.475\textwidth]{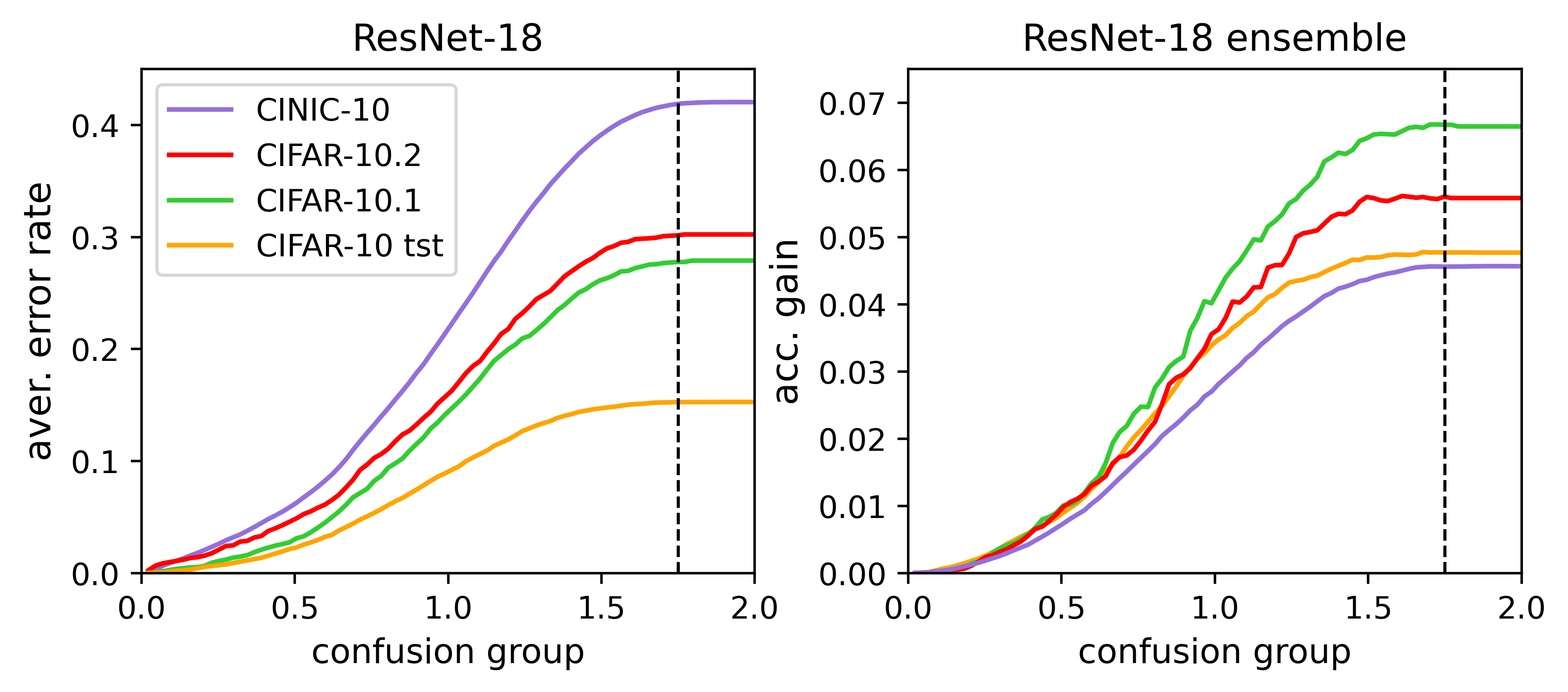}
   \includegraphics[width=0.475\textwidth]{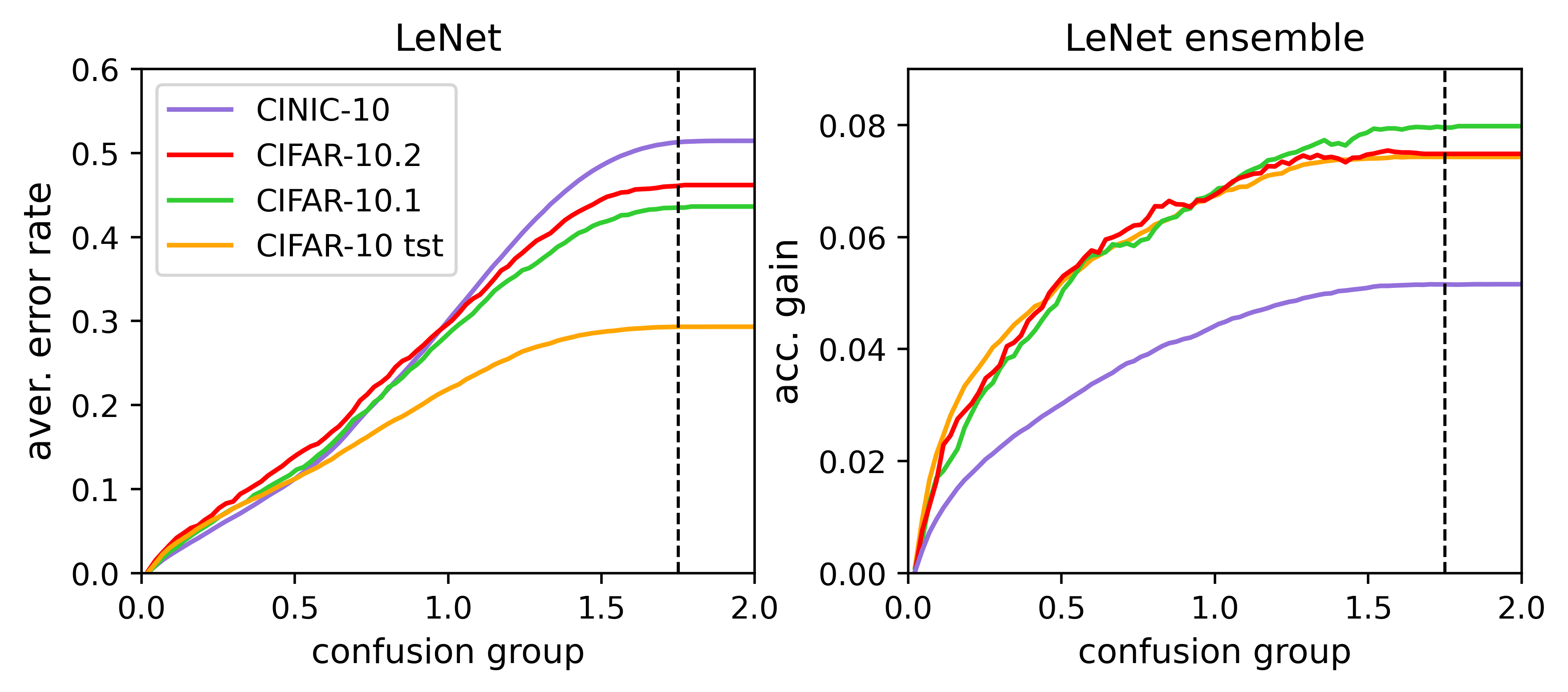}
   \includegraphics[width=0.475\textwidth]{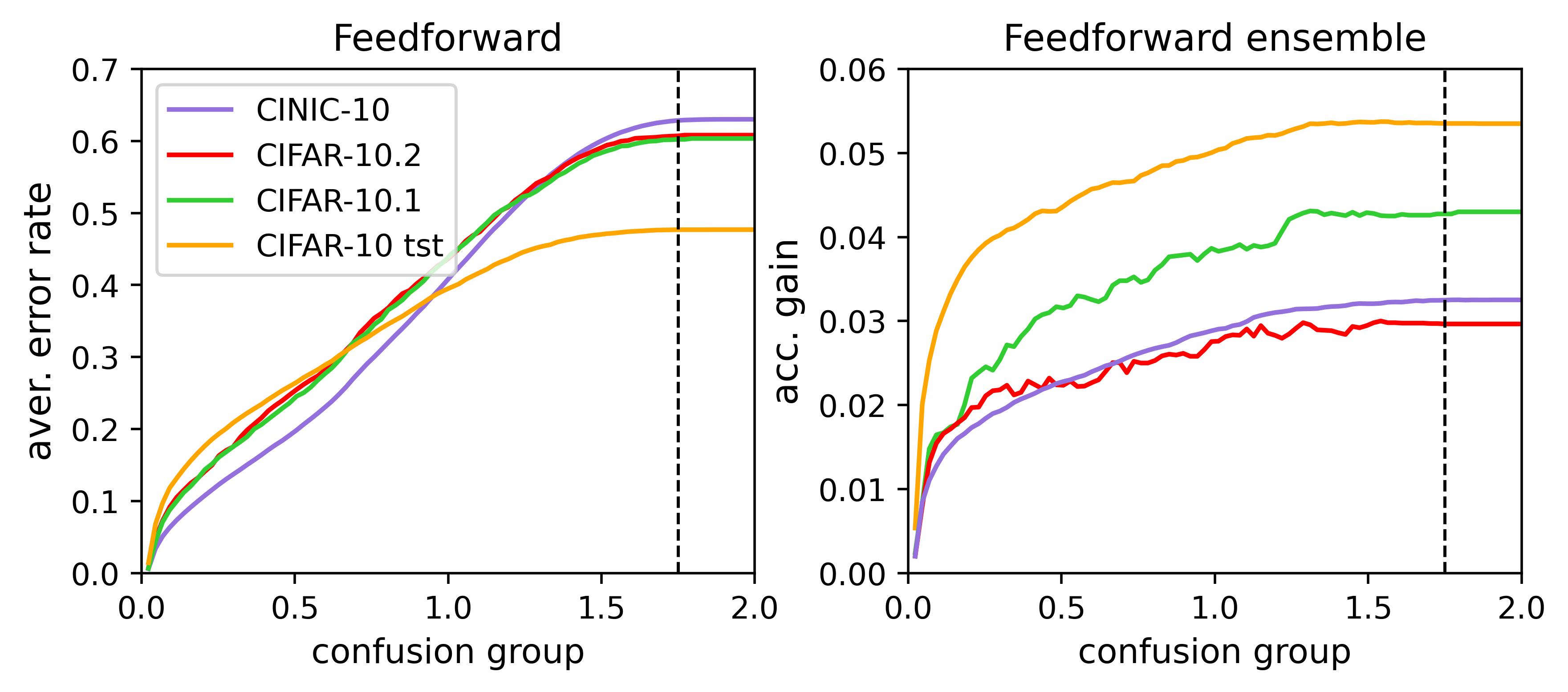}
   \end{center}
%   \vspace{-0.3cm}
  \caption{\label{fig:error-rates} \textbf{Left panels: The majority of the accuracy drop in the OOD datasets comes from high confusion score samples.} Average number of mistakes over the samples for three model architectures: ResNet-18, LeNet, and feedforward networks over 10 seeds each:
  We observe that the average error rates for the OOD datasets are close to the ID dataset for samples with low confusion group samples and the majority of the deviation comes from the high confusion groups.
  \textbf{Right panels: Accuracy gain due to ensembling.} Cumulative difference between the accuracy of the ensemble and average accuracy: we observe that (i) for ResNet18, the accuracy gain comes mainly from high confusion groups, (ii) for LeNet, the accuracy gain is distributed among high confusion and low confusion groups, (iii) for the feedforward net, most of the accuracy gain comes from low confusion groups.}
\end{figure}

\section{What causes the OOD accuracy drops?}
\label{sec:acc-drop}

We established how the confusion scores are calculated to quantify the difficulty of test images in Section~\ref{sec:foundations}. We have seen in Fig.~\ref{fig:confusion-dists} that the average confusion score is the highest in CINIC-10 which is known to be noisiest in comparison to the other two OOD datasets, and the lowest in CIFAR-10. Next, by partitioning the test datasets according to their confusion scores, we identify what makes the OOD datasets harder than the ID dataset.

\textbf{Experimental Setup:} For ResNet-18s, we used a learning rate of $0.01$, training for $200$ epochs. For LeNets, we used a learning rate of $0.001$, training for $250$ epochs, scaling the number of filters in the first two layers as $(w, 2w)$ for $w=\{2, 4, \ldots, 128\}$ where $w=16$ corresponds to the original LeNet architecture. For feedforward nets, we trained $2$ and $3$ layers architectures, scaling the width as $w=\{2, 4, \ldots, 1028\}$, using a learning rate of $0.001$, training for $1000$ epochs. For all architectures, we used a batch size of $128$, the Adam optimizer with default Pytorch parameters, and an early stopping criterion based on the minimum test accuracy (ID) during training. We trained $10$ random seeds for each architecture (i.e. changing initialization and the order of training samples). No data augmentation was used to study the plain structure of the training dataset, except for the normalization of the color channels.

\subsection{Behavior of different models on confusion groups}

We first observe that - although the participation ratio of the low confusion score groups is very high - the average error rate in this subpopulation is low compared to the cumulative error rate of the model (equivalently test error) for ResNet and LeNet. This is expected, as these models have good inductive biases due to the convolutional filters in the architecture.
In contrast, for the feedforward network we see that the error rate is distributed more homogeneously over confusion groups. This indicates that the feedforward network does not extract the simple features/correlations that enable almost perfect accuracy on low confusion groups.

However, for the high confusion groups, the error rates of all the networks are more similar, indicating that the quality of the simple modalities extracted does not dramatically affect the error rates in this regime. This may be due the absence of simple features and class-specific spurious correlations in high confusion groups.

\section{Predicting OOD accuracies using the confusion scores}
\label{sec:pred-ood-accs}

\begin{figure*}[t!]
   \begin{center}
   \includegraphics[width=0.8\textwidth]{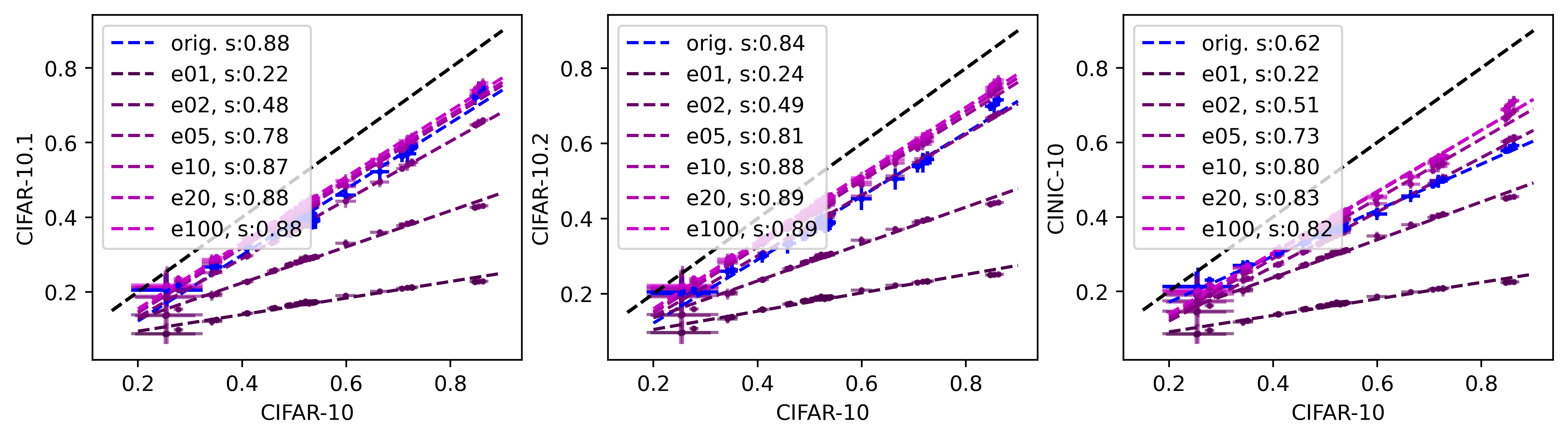}
   \end{center}
  \vspace{-0.5cm}
  \caption{\label{fig:OOD-acc-prediction} \textbf{Predicting the OOD accuracies from the ID accuracies using the confusion scores.} Blue: The mean accuracies of $300$ feedforward networks with depths $[2,3,4]$, with widths $[2,4,8, \ldots, 1024]$ at each layer, and $70$ LeNets with $2$ convolutional layers with filter numbers $[2, 4, \ldots, 128]$ in the first layer and doubled number of filters in the second layer, followed by three feedforward layers, and $40$ ResNets (ResNet18, ResNet34, ResNet50, and ResNet101); each architecture has $10$ seeds. Horizontal and vertical lines represent $\pm2$ standard deviations calculated for the fixed architectures across $10$ runs. Purple dots represent the predictions of OOD accuracies from the ID accuracies using the subpopulation model based on confusion scores.
  In the legend, $e_t$ represents the predictions based on the partitioning based on the entropy scores of epoch $t$.
  For all epochs, we predict a linear relationship between ID and OOD accuracies, underestimating for very early epochs and overestimating in later epochs.}
\end{figure*}

Given an ensemble of trained architectures (i.e. ResNets), we bin the confusion scores into $n$ levels, i.e. samples with confusion scores $[0, \log(C)/n]$ are collected in the lowest confusion group. We then calculate the ID accuracies in every confusion group. Finally, we calculate the weighted average of accuracies per confusion group according to the population ratios of the OOD datasets which gives a prediction of the average accuracy. In Figure~\ref{fig:OOD-acc-prediction}, we used $ n = 40$.

The assumption here is that the samples in the same confusion bucket are \textit{identically distributed} with the probability density function $p_i(x)$, even if they come from different test datasets. Under this assumption, we model the test distributions as mixtures
\begin{align*}
    &p^{\text{ID}}(x) = \sum_{i=1}^{K} \alpha_i p_i(x) \\
    &p^{\text{OOD}_{j}}(x) = \sum_{i=1}^{K} \beta_i^{j} p_i(x)
\end{align*}
where $\sum_{i=1}^{K} \alpha_i = 1$ and $\sum_{i=1}^{K} \beta_i^j = 1$ for $j=\{1,2,3\}$ representing CIFAR-10.1, CIFAR-10.2, and CINIC-10 respectively. We will refer $\alpha_i$ and $\beta_i^{j}$ as participation ratios.

We argue that the difference between ID and OOD datasets can be partially explained via participation ratios of identically distributed subpopulations. In yet another but identical formulation, we assume that the \textit{distribution shift} in the OOD datasets is mainly due to the change in the participation ratios. In particular

\begin{itemize}
    \item We can explain the big part of the accuracy drop that comes from the shift in the distribution of confusion scores in the OOD datasets.
    \item However, our method does not take into account the accuracy drop due to class-specific spurious correlations or label corruptions.\footnote{This is a fundamental limitation which requires either context understanding or manual inspection.}
\end{itemize}

We find that the prediction works the best for the CIFAR-10.1 OOD dataset which indicates that CIFAR-10.1 dataset does not suffer from spurious correlations or label corruption too much.
On the other other hand, we find that our prediction is the worst for the CINIC-10 dataset since it has the largest amount label corruption \cite{darlow2018cinic}.
This is consistent with our observation in the low confusion group, the error rates in the CINIC-10 is the highest and that of CIFAR-10.1 is the lowest (see Figure~\ref{fig:error-rates}).

We find that using only the early phase of training, our predictions underestimate the OOD accuracies whereas using only the late phase of training, our predictions overestimate the OOD accuracies (see Fig.~\ref{fig:OOD-acc-prediction}).
We underestimate the OOD accuracies early in training since we assign high confusion scores even to easy images. Looking into the transition from early to late training, we observe that the end of Phase-I (i.e. epoch $t=5$ in our experiments) predicts the OOD accuracies the best! This may be linked to the point where the network stops learning features and starts memorizing individual examples (i.e., test images that are not similar to any of the examples in the training dataset, these can also be viewed as a minority group without cluster formation).

The authors of \cite{jiang2021assessing} find that the disagreement rate of two independent runs of training on the same dataset predicts the test error surprisingly well.
However, agreement-based methods do not take into account the cases where two models agree, but on the wrong class (either due to spurious correlations, or label corruption). Therefore, we would expect poor prediction performance of this model for predicting CINIC-10 accuracy where many data samples are corrupted. The authors of \cite{vedantam2021empirical} employ a large variety of methods for predicting the generalization performance on OOD datasets and report that the ID accuracies and the mean entropy scores of the datasets predicted by trained networks correlate the best with the OOD accuracies. Our prediction method does not only correlate with the OOD accuracies, but also outputs a good estimation (see Fig.~\ref{fig:OOD-acc-prediction}). The precision of our OOD accuracy predictor comes from the partitioning of the test datasets instead of relying on average metrics.

\section{Interpretations}
\label{sec:interpretations}

\begin{figure}[h!]
   \begin{center}
   \includegraphics[width=0.2\textwidth]{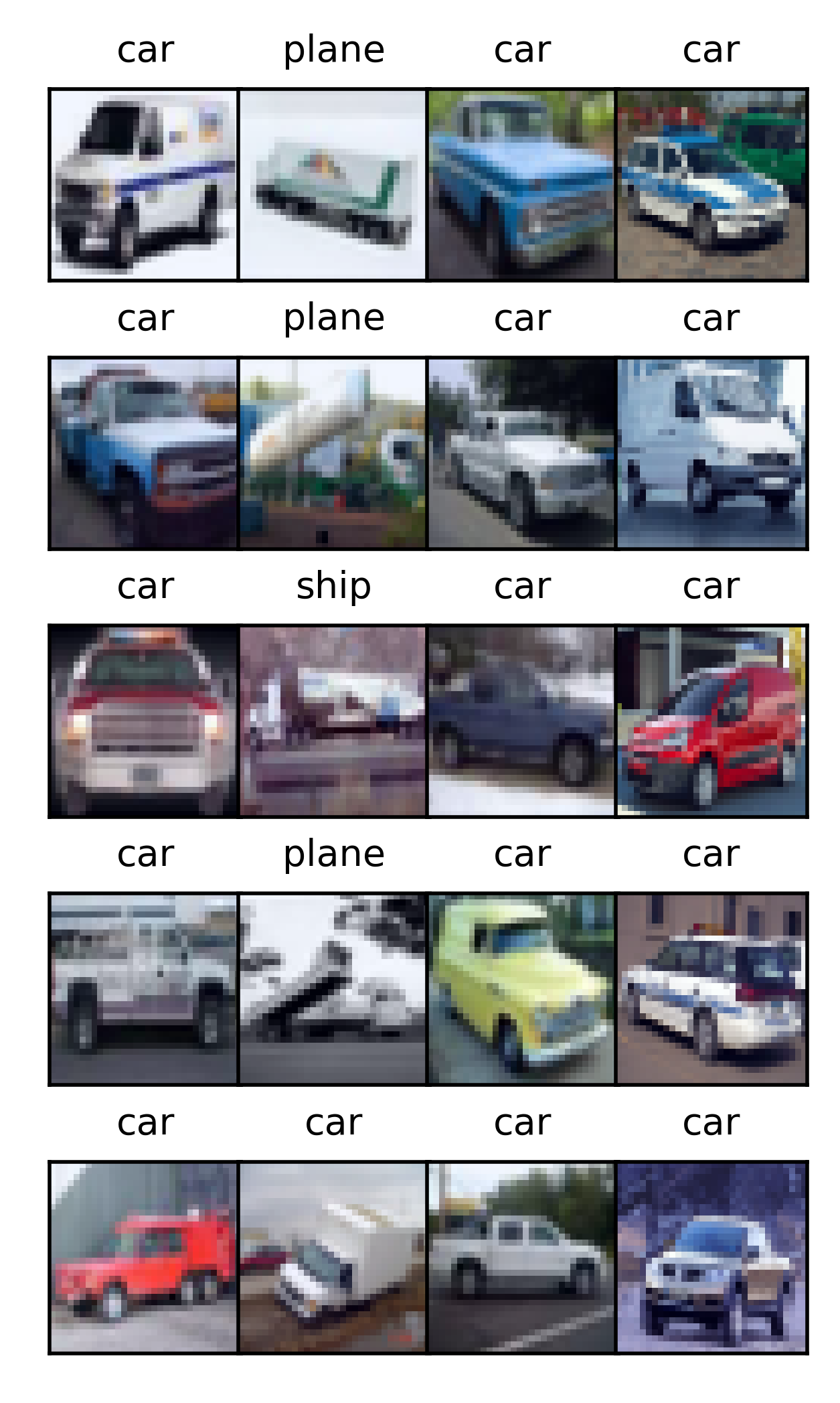} \hspace{0.1cm}
    \includegraphics[width=0.2\textwidth]{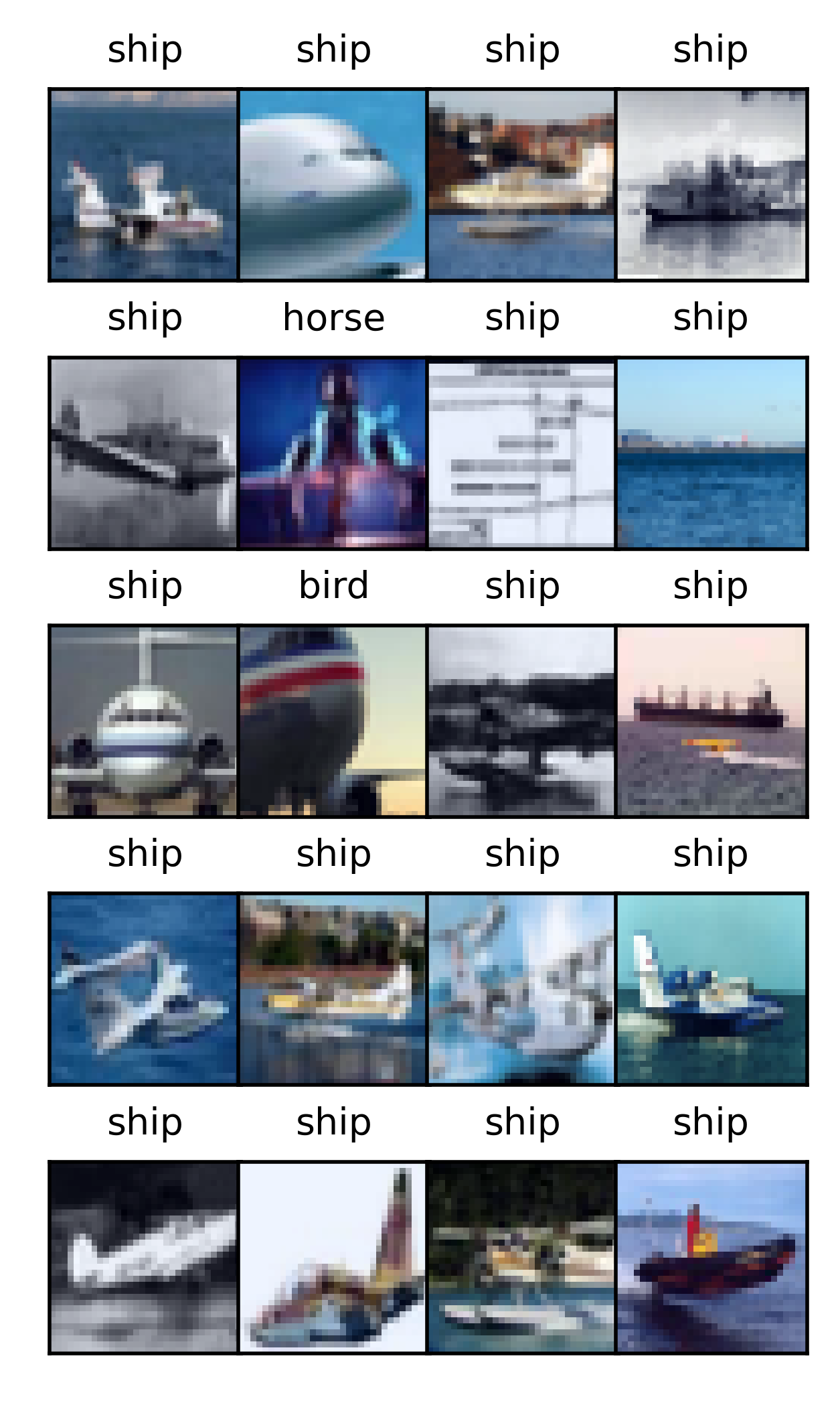}
    \hspace{0.1cm}
    \includegraphics[width=0.2\textwidth]{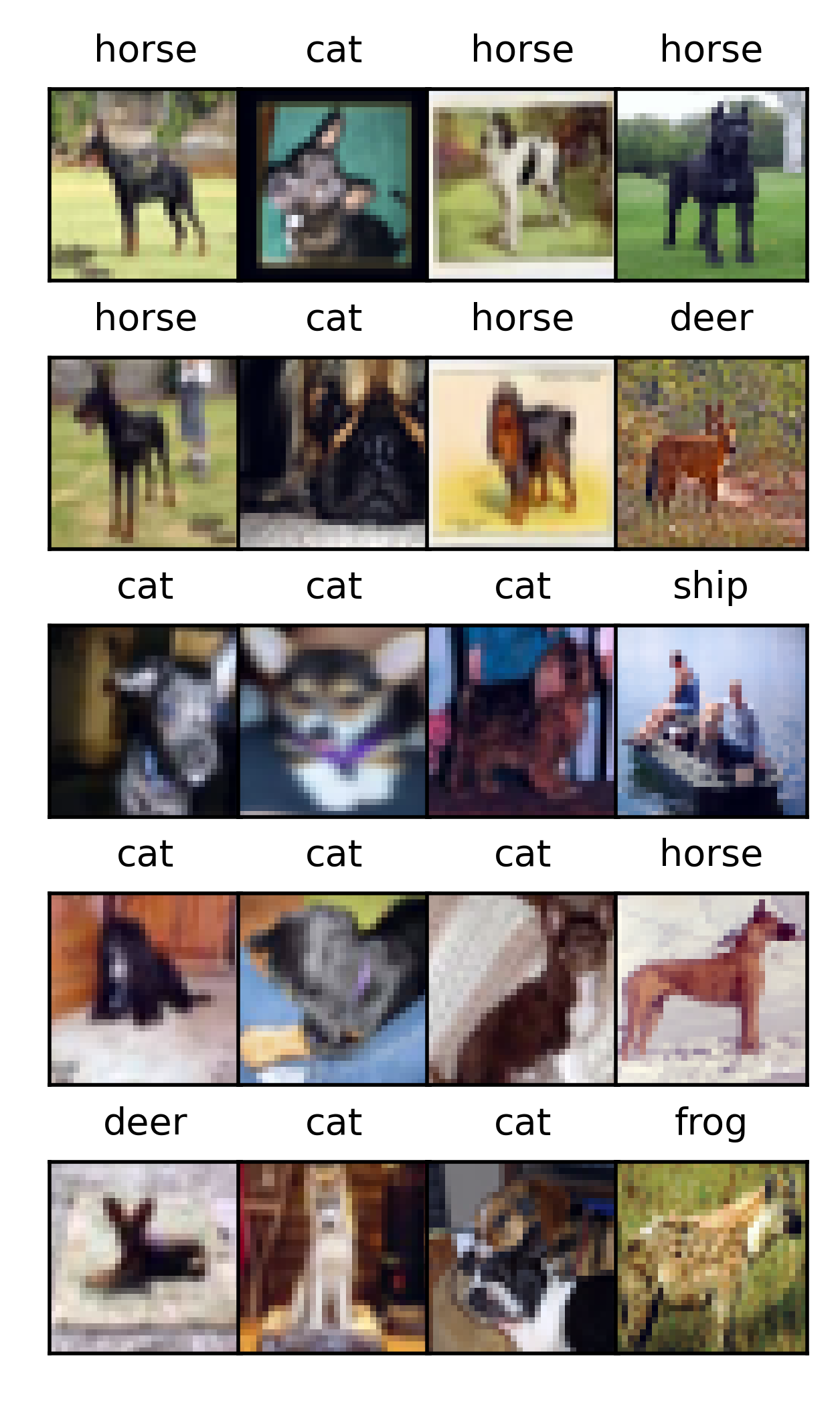}
    \hspace{0.1cm}
    \includegraphics[width=0.2\textwidth]{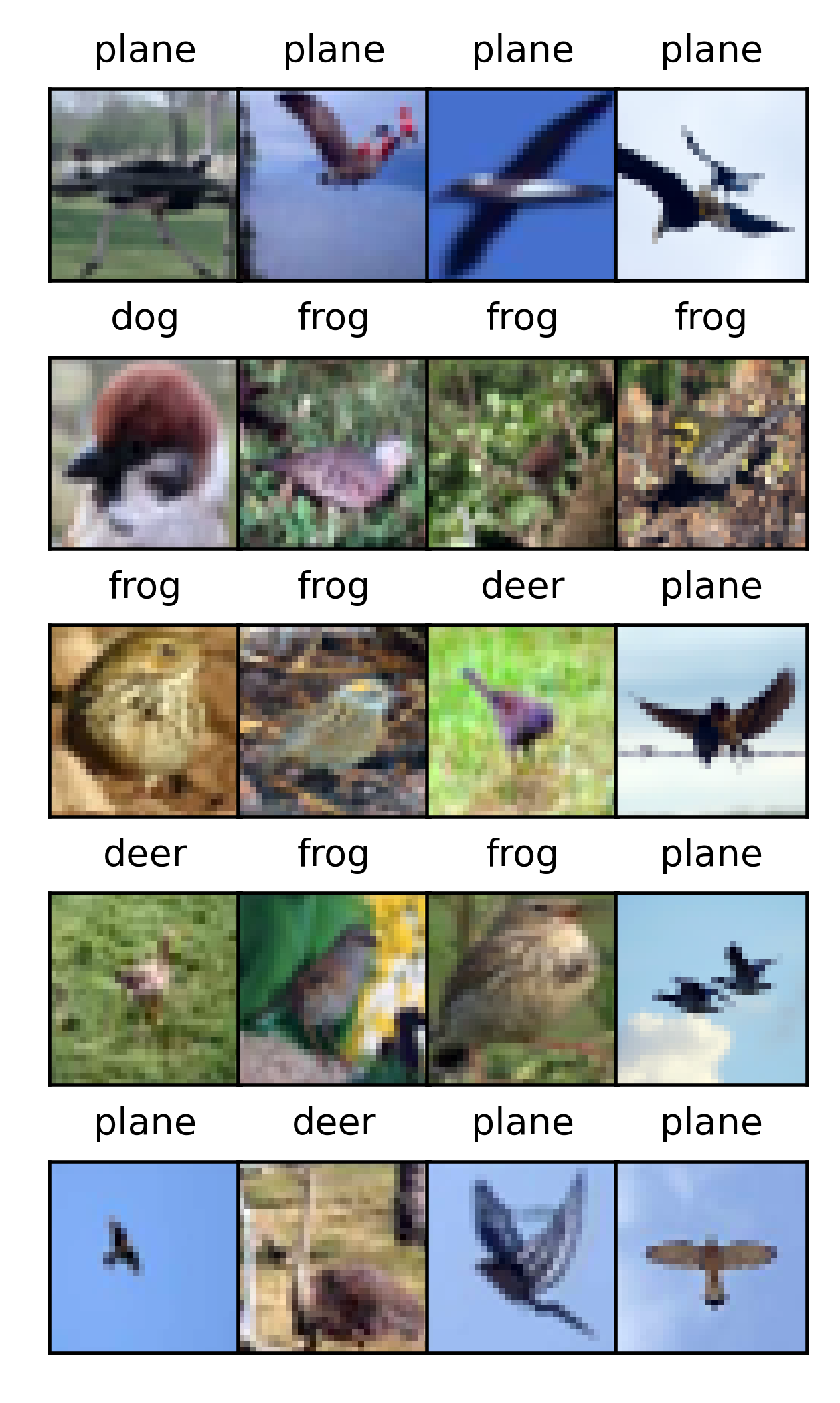} \\
    \vspace{-0.25cm}
    {\scriptsize \textbf{(a)} Truck \hspace{2.25cm} \textbf{(b)} Plane \hspace{2.25cm} \textbf{(c)} Dog \hspace{2.25cm} \textbf{(d)} Bird}
   \end{center}
  \vspace{-0.3cm}
  \caption{\label{fig:sp-corr} \textbf{Lowest confusion score mistakes from the classes of truck, plane, dog, and bird from left to right}. Titles represent the consensus prediction of the bag of resnets.
  In each class, the first column represents images from CIFAR-10, the second from CIFAR-10.1, the third from CIFAR-10.2, and the last from CINIC-10. \textit{(a) Truck label:} numerous examples of label corruption, many of low confusion truck mistakes are in fact cars. \textit{(b) Plane label:} round planes without wings, and planes without distinguishing plane features with a sea background are classified as ships (spurious correlation). \textit{(c) Dog label:} dogs with pointy ears are classified as cats since this is the distinguishing feature of the cat class in the CIFAR-10 training data. \textit{(d) Bird label:} birds with open and sharp wings flying over a blue sky are classified as planes. Finally both in the bird and dog classes, we observe that the grassy and green backgrounds are attributed to horse and deer labels respectively when distinguishing class features are not present in the images.
  }
\end{figure}

Note that the agreement based measures such as in \cite{jiang2021assessing} fall short in separating the easy images from the images that include spurious correlations. We argue that low confusion group mistakes are due to spurious correlation or label corruptions and high confusion group mistakes are due to weak spurious correlations.
The usual spurious correlation datasets only have two classes (i.e., binary classification setup). With our confusion scores, we see that one class is not only spuriously correlated with exactly one other class but with many, in fact (best seen in the bird class-appendix).
In this section, we give an interpretation of how these correlations are distributed over the confusion group subpopulations.

\noindent \textbf{Low confusion score samples:} (Class-specific spurious correlations \& Label corruptions \& Easy-Typical) Typically samples with low confusion scores achieve low entropy scores/high confidences very early on in training. Similarly, due to the simultaneous extraction of simple features and correlation modalities early in training, the samples exhibiting spurious correlations are also assigned low confusion scores. Here is a list of samples exhibiting various forms of class-specific spurious correlations from Fig.~\ref{fig:sp-corr}: (i) planes with a sea background (spuriously correlated with the ship label), (ii) dogs with pointy ears (spuriously correlated with the cat label), (iii) birds with open wings and blue sky (spuriously correlated with the plane label), and (iv) birds with earthy background, green/brown color distribution, and dotty texture (spuriously correlated with the frog label).
Finally, we capture some images with label corruption in low confusion score groups since the majority of the networks make the same prediction early in training on these samples (see the car images in the truck class in Fig~\ref{fig:sp-corr}).
For these samples, the entropy scores do not increase later in training.

Note that achieving $0$ entropy is identical to achieving confidence $1$, thus the low confusion samples can also be identified from their confidence during training.
Moreover, the high similarity of the histograms of the entropy scores at epoch $t=10$ (Fig.~\ref{fig:time-eval-entrop-scores}) and the confusion scores (Fig.~\ref{fig:confusion-dists}) suggests that the low confusion score samples can potentially be identified using only early training dynamics (Phase-I and II).

\noindent \textbf{High confusion score samples:} (Weak spurious correlations \& missing class-specific features/spurious correlations) An ensemble of networks make conflicting predictions on these images, consistently during training. Therefore neither the simple features nor simple correlations (such as class-specific spurious correlations) that are picked up consistently by different seeds are not present in these images. Ruling out the possibilities of simple modalities in high confusion score samples, there is an abundance of freedom in the input space deviating from the simple modalities present in the training dataset. Some modalities we identified in high-confusion samples are weak-spurious correlations such as gray-scale and white-background.

\section{Conclusion}
We present a label-free method of evaluating how confusing a given sample is to a given model. This measure highlights several aspects of input-output correlations which are also intimately linked to how OOD a given test dataset is. Furthermore we show that the proposed measure can be used to estimate the OOD accuracies.

The confusion score is a measure that depends on the data as well as the chosen model size and architecture. Such dependence is essential as data-centric perspectives disregard potential capacities of the model, and model-centric perspectives disregard the structure of the data. This work itself is limited in its applicability: our observations are expected to hold under the assumption of consistent labeling and sampling for the OOD dataset at hand. A thorough understanding of the structure of data necessarily requires understanding the context of the dataset, as well as understanding how data is collected and labeled. We believe that the approach outlined in this paper may help guide investigations of datasets and open new ways to see data-model relationship.

\section*{Acknowledgements}
We thank David Lopez-Paz, Diane Bouchacourt, Mark Ibrahim, Pascal Vincent, Stéphane d'Ascoli, Leon Bottou, Kamalika Chaudhuri, Stéphane Deny, Mohammad Pezeshki, Maksym Andriushchenko, Robert Geirhos, and Mohamed Ishmael Belghazi for many great discussions and feedback.

\bibliographystyle{unsrt}
\bibliography{references}

\newpage
\appendix

\section{Dynamics of learning from the dataset point of view}

In this section, we will dive deep into the training dynamics of neural networks in order to understand what features and correlations are picked up during three phases of training, what is the main difference between ID and OOD datasets, and a more detailed analysis of our confusion score.

\subsection{Average metrics during training}

For completeness, we will present the average accuracies and the average confidences during training for the test datasets and the training datasets as a completementary to Fig~\ref{fig:time-eval-entrop-scores} in the main. However note that the average metrics are dominated by the majority populations therefore disguising what happens for minority groups or isolated test points in the test datasets. Therefore in the following subsections, we will focus on individual data points or subpopulations, going beyond the standard analaysis with average metrics.

\begin{figure*}[h!]
   \begin{center}
  \includegraphics[width=0.3\textwidth]{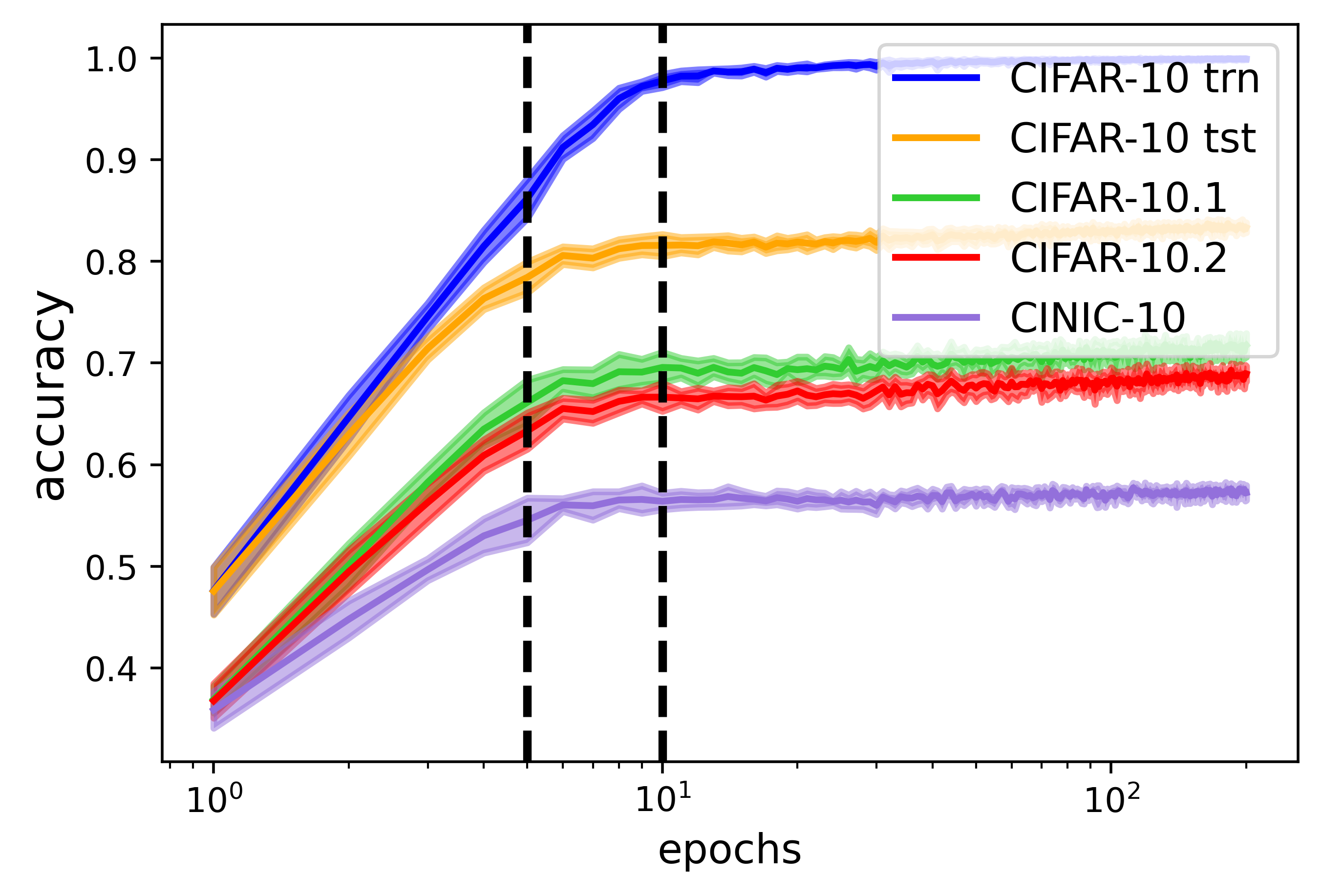} \hspace{1cm}
  \includegraphics[width=0.3\textwidth]{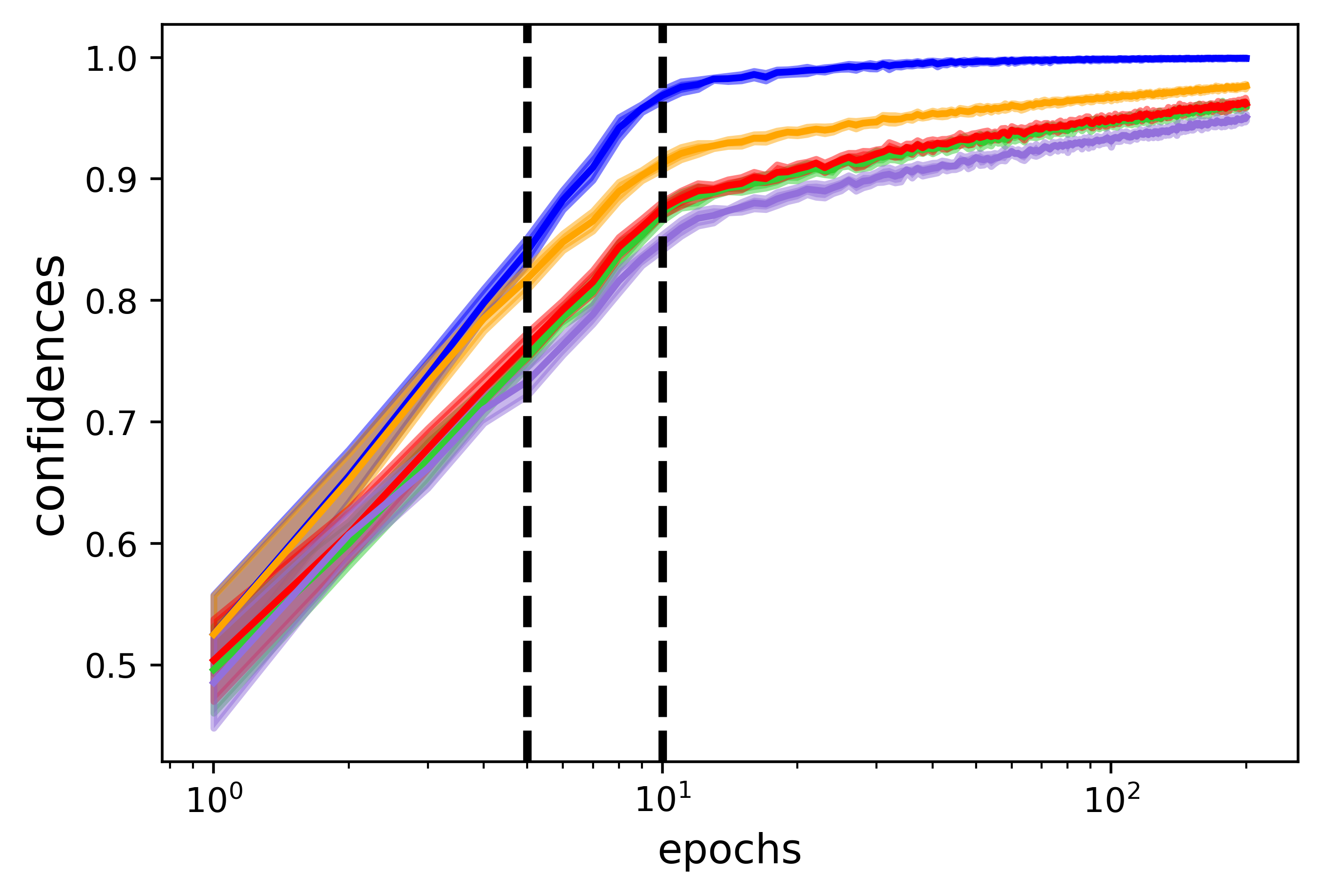}
   \end{center}
  \vspace{-0.3cm}
  \caption{\label{fig-app:scores-during-training} \textbf{The evolution of the accuracies and aver. confidences over training}. The optimal stopping based on test accuracies are different for all $4$ test datasets: $t=170, 120$, $84$, $57$ respectively for CIFAR-10 test, CIFAR-10.1, CIFAR-10.2, CINIC-10 (for one of the seeds) which challenges the common practice of choosing the early stopping time based on the test accuracy of the ID dataset.}
\end{figure*}

\textbf{Entropy scores $s_t(x)$ is not stable during training, especially in Phase-III}
\begin{figure*}[h!]
   \begin{center}
  \includegraphics[width=0.7\textwidth]{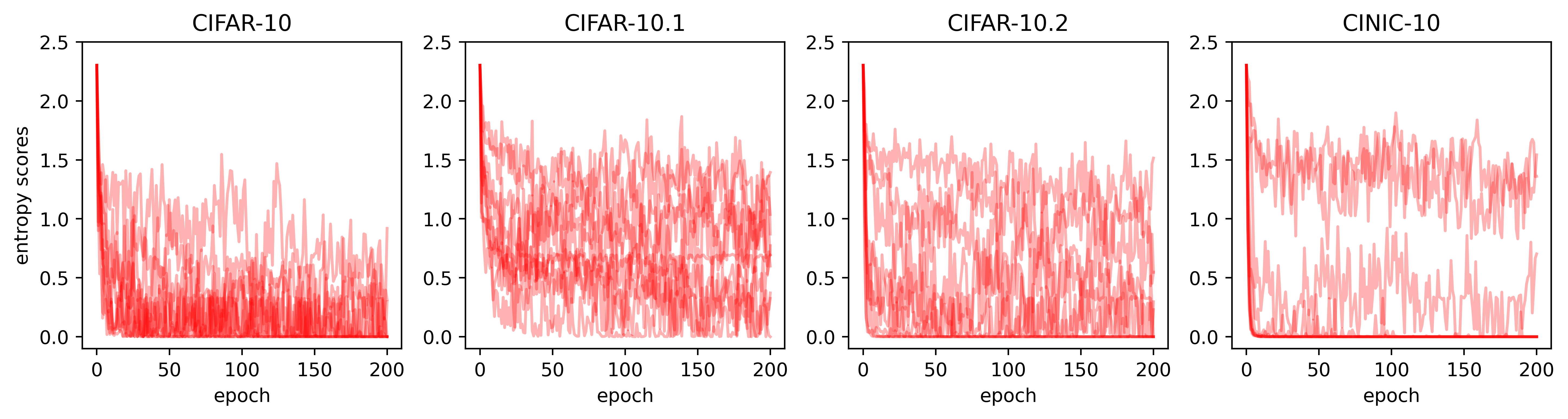} \\
  \vspace{0.1cm}
  \includegraphics[width=0.7\textwidth]{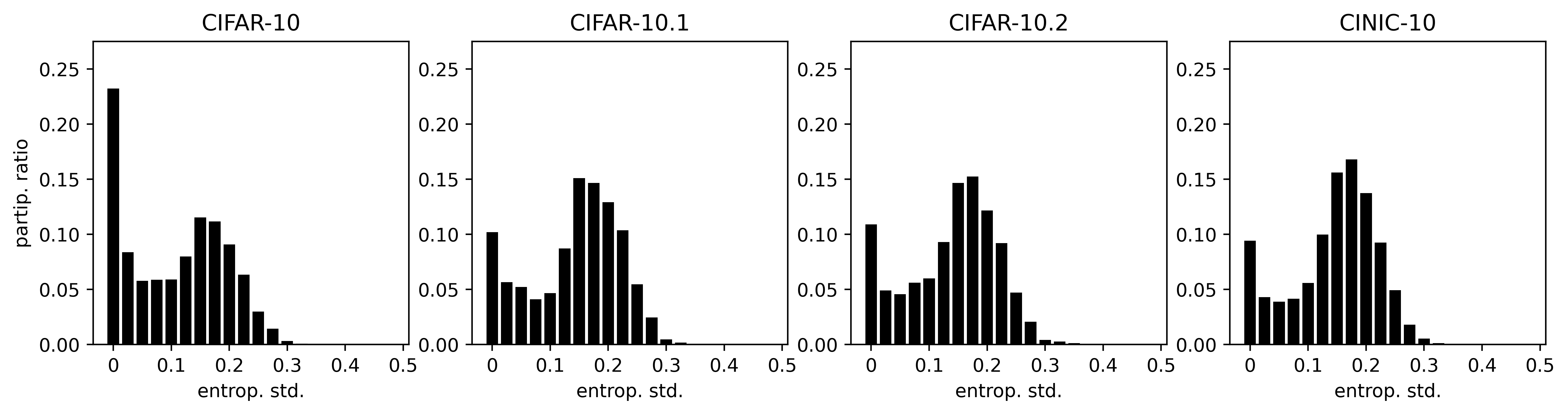}
   \end{center}
  \vspace{-0.3cm}
  \caption{\label{fig-app:entropy-instable-training} \textbf{Top: Entropy score trajectories of randomly selected test images}. We observe that the entropy scores during training wildly fluctuates. One exception is for the low entropy samples: these are learnt early in the training and their class probability distribution during training remains stable. We observe the same behavior for tens of different random sampling, however only $10$ samples from each dataset are plotted for the visibility of single trajectories. \textbf{Bottom: The histogram of standard deviation of entropy scores during the phase-III of training (after epoch $10$).} We observe that the majority of the test samples in the datasets we investigated exbihit high fluctuation of the entropy scores during training. The means of the standard deviations are $0.112$, $0.148$, $0.145$, and $0.152$, respectively. We observed quantitatively very similar histograms for the standard deviations of entropy scores after epoch $50$.}
\end{figure*}

\subsection{Time evolution of the entropy scores}

We presented the evolution of the entropy scores for CIFAR-10 test dataset in the main Fig.~\ref{fig:time-eval-entrop-scores}. Below we present evolutions for CIFAR-10 training, CIFAR-10.1, CIFAR-10.2, and CINIC-10 datasets.

\begin{figure}[h!]
   \begin{center}
    \includegraphics[width=0.48\textwidth]{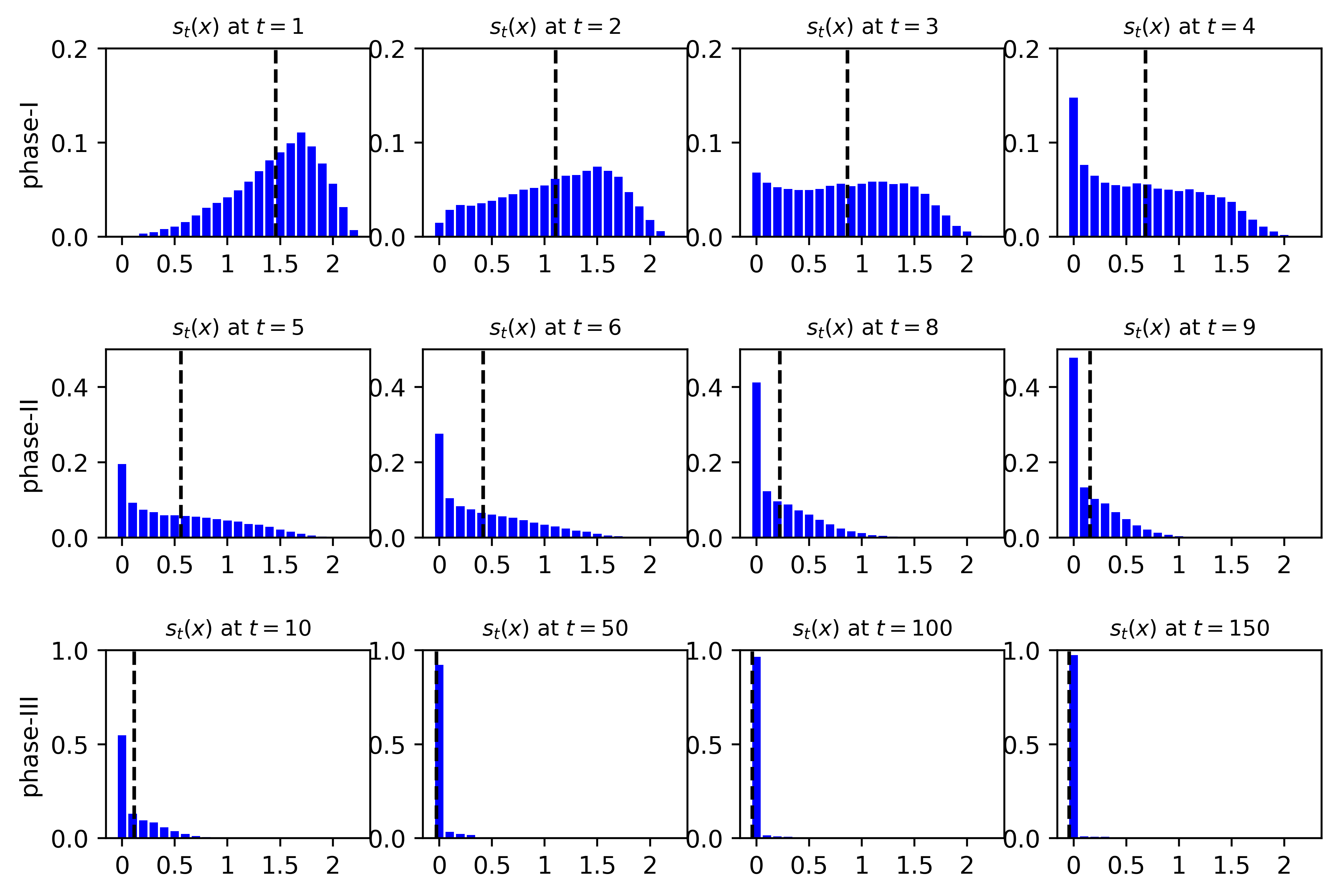} \hspace{0.2cm} \includegraphics[width=0.48\textwidth]{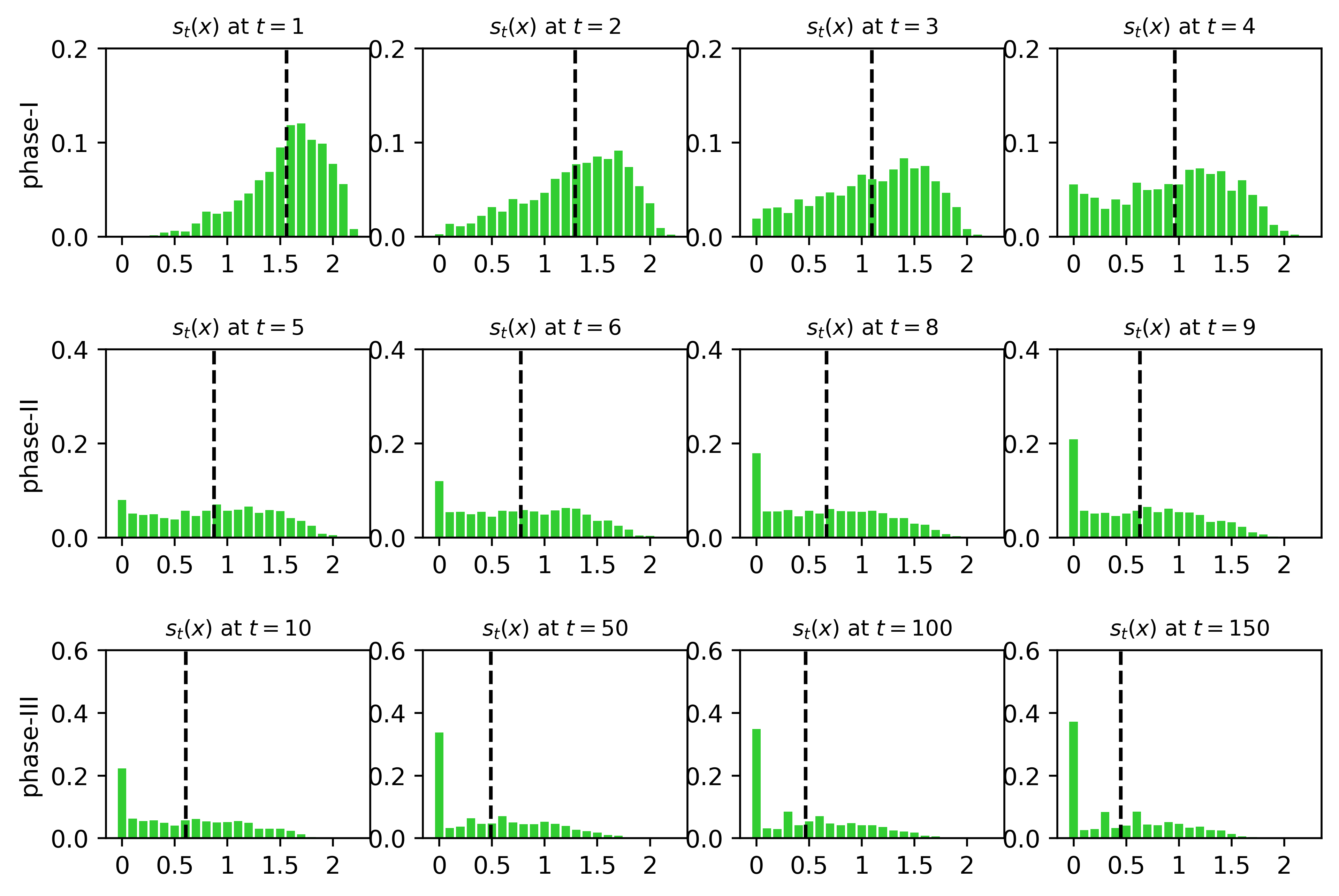} \\
    {\scriptsize \textbf{(a)} CIFAR-10 train \hspace{5cm} \textbf{(b)} CIFAR-10.1} \\
    \includegraphics[width=0.48\textwidth]{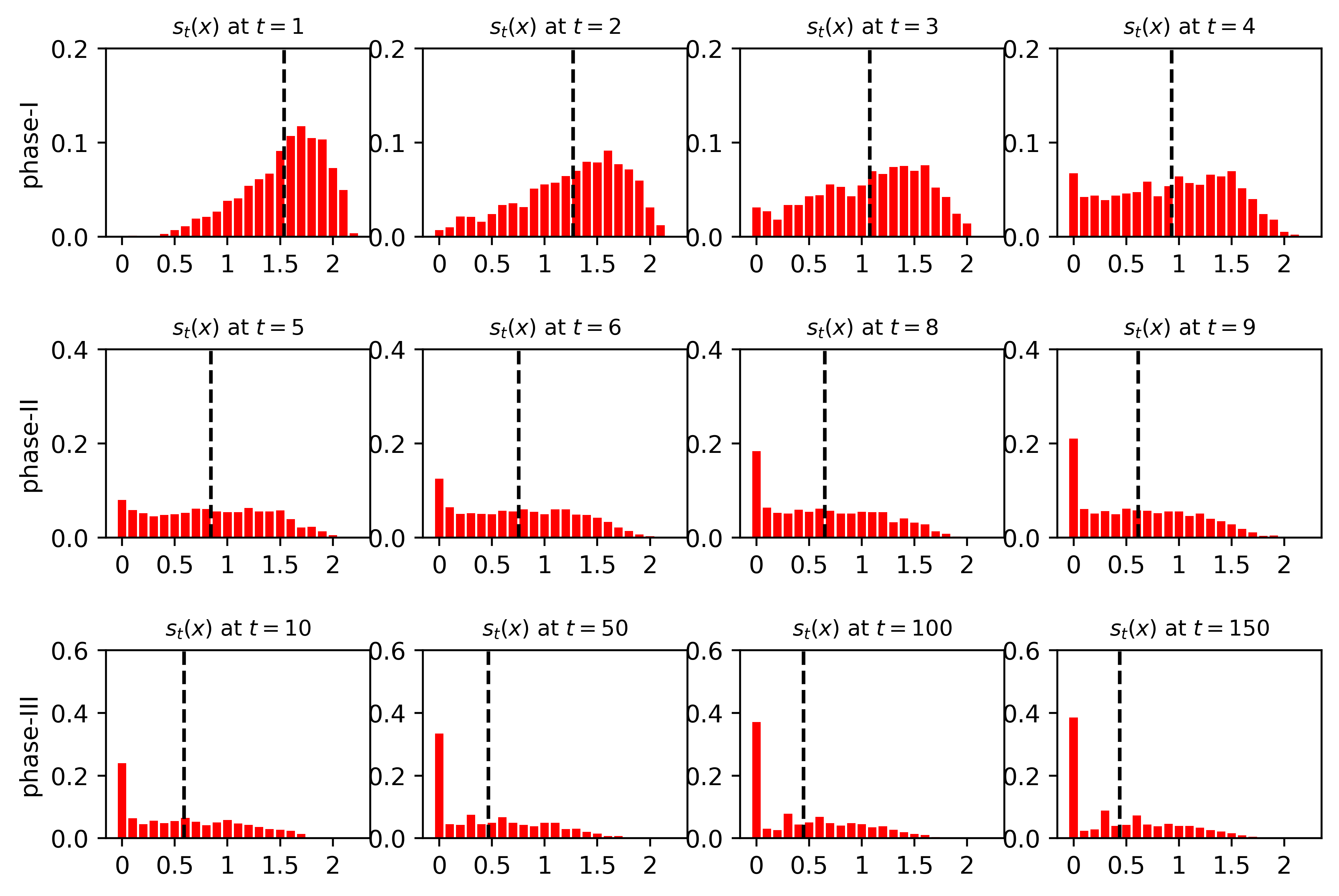} \hspace{0.2cm} \includegraphics[width=0.48\textwidth]{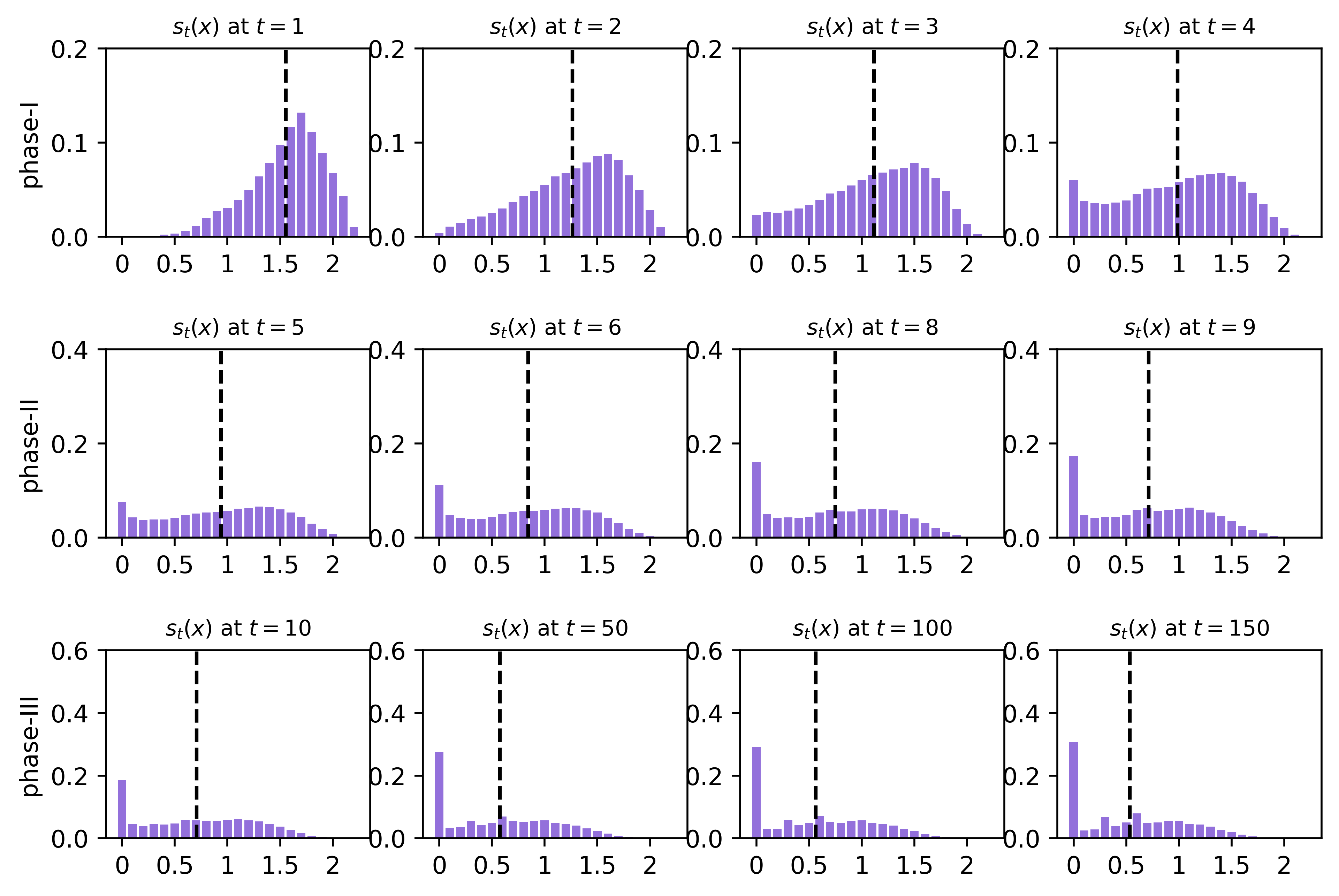} \\
    {\scriptsize \textbf{(c)} CIFAR-10.2 \hspace{5cm} \textbf{(d)} CINIC-10} \\
   \end{center}
   \vspace{-0.3cm}
  \caption{The participation ratio of the low entropy groups between datasets are similar. However due to softmax, in the training dataset, the confidences in this group are pushed up high which in turn results in bucketing all the samples, both easy and in fact possibly highly confusing in the same bucket.}
\end{figure}

\pagebreak

\section{Accuracy across confusion groups for various model families}

\begin{figure}[h!]
  \begin{center}
    \includegraphics[width=0.9\textwidth]{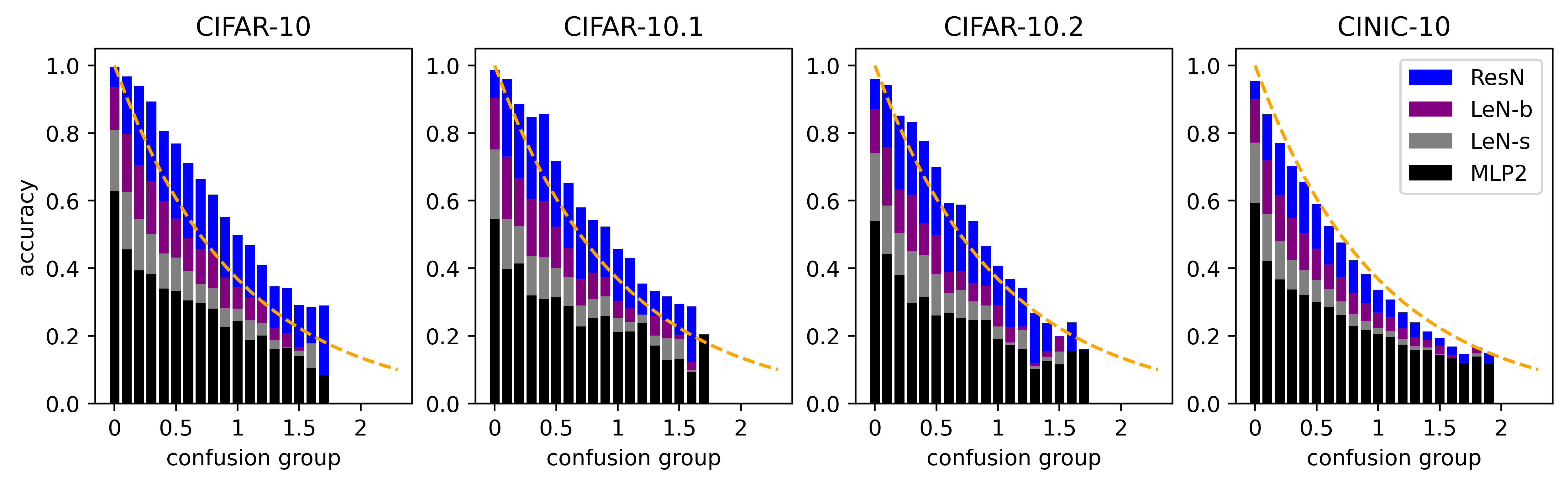}
  \end{center}
  \vspace{-0.3cm}
  \caption{\label{fig:appendix-acc-per-group} The average accuracies of (i) ResNets (10 resnet18s, 10 resnet34s, 10 resnet50s, and 10 resnet101s), (ii) big LeNets with the number of filters $w \in \{16, 32, 64, 128\}$ in the first layer and doubling the second layer, (iii) small LeNets with the number of filters $w \in \{2, 4, 8\}$ in the first layer and doubling the second layer, and (iv) two-layers networks with widths $\{2,4,8,\ldots, 1024\}$. Decreasing model complexity hurts accuracies across subgroups in a structured way, i.e. drop in accuracy in each subgroup is homogenous. The orange line represents $ e^{-x}$  where $ x $ is the confusion score. Preliminary observations suggest that the accuracy levels per group can be modeled with $ C^{(\alpha -1)} e^{-\alpha x}$ where $\alpha\in (0,1)$ represents the model complexity ($C$ is the number of classes): for $\alpha=1$ we have the orange line and where for $\alpha = 0$ we have the random model which is guessing randomly the correct class for each subgroup.}
 \end{figure}

\section{Overparameterization and Ensembling via Label-Dependent Partitioning}

In this section, we use a label-dependent procedure of partitioning the test datasets into three groups. We take $70$ LeNets with architectures $(w, 2w)$ filters in the first two layers with $w=\{2, 4, \ldots, 128\}$ and $10$ seed for each architecture. For each test datapoint, we count how many of $70$ networks classify the point correctly. We collect the test images that are correctly classified by more than $2*70/3$ networks in the easy subpopulation, those that are correctly classified by less than $70/3$ networks in the hard population, and the other ones in the medium subpopulation. We report the participation ratios for these three subpopulations in Table~\ref{table:ratios-app}. Note that the easy subpopulation makes up the biggest proportion typically which suggests that this is also the majority subpopulation. For the hard subpopulation, the accuracies fluctuate around the chance level ($0.1$ for $10$ classes), that is why we also call this population ambigious. Note that for CINIC-10 dataset, the ambigious population is significantly bigger than the other two OOD datasets.
\begin{table}[h!]
\caption{Participation ratios of subpopulations \label{table:ratios-app}}
\label{sample-table}
\begin{center}
\begin{small}
\begin{sc}
\begin{tabular}{lcccr}
\toprule
Dataset & Esay & Medium & Hard \\
\midrule
CIFAR-10 tst & 0.605 & 0.210 & 0.185 \\
CIFAR-10.1 & 0.413 & 0.266 & 0.320 \\
CIFAR-10.2 & 0.403 & 0.248 & 0.350 \\
CINIC-10 & 0.358 & 0.211 & 0.430 \\
\bottomrule
\end{tabular}
\end{sc}
\end{small}
\end{center}
\end{table}

\begin{figure}[t!]
    \centering
    \includegraphics[width=0.7\textwidth]{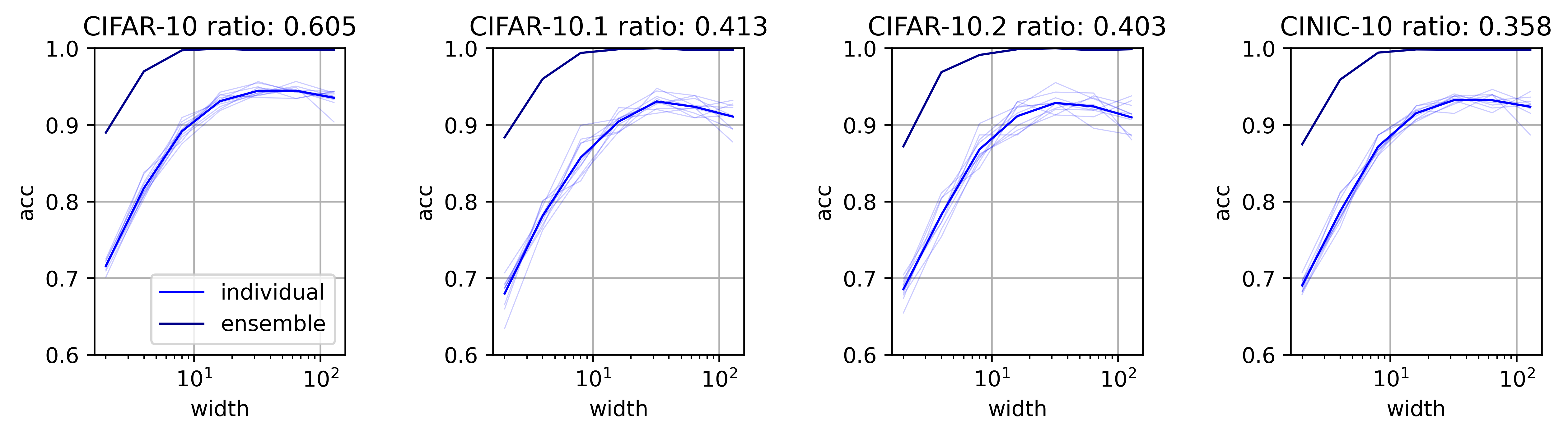} \\
    \vspace{-0.2cm}
    {\scriptsize \textbf{(a)} Majority/Easy subpopulation} \\
    \includegraphics[width=0.7\textwidth]{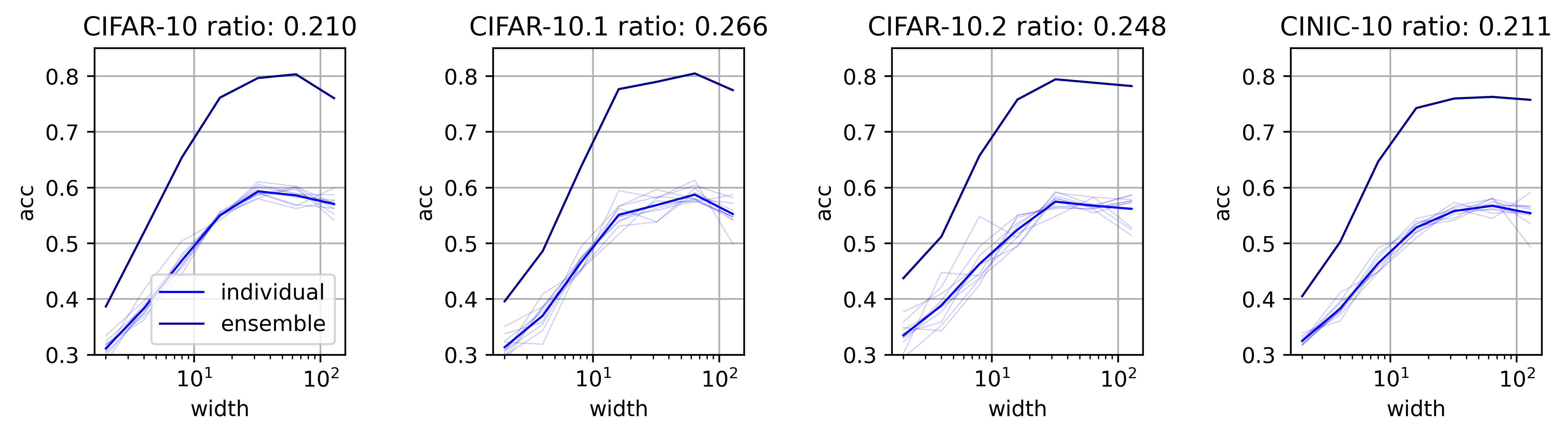} \\
    \vspace{-0.2cm}
    {\scriptsize \textbf{(b)} Medium subpopulation} \\
    \includegraphics[width=0.7\textwidth]{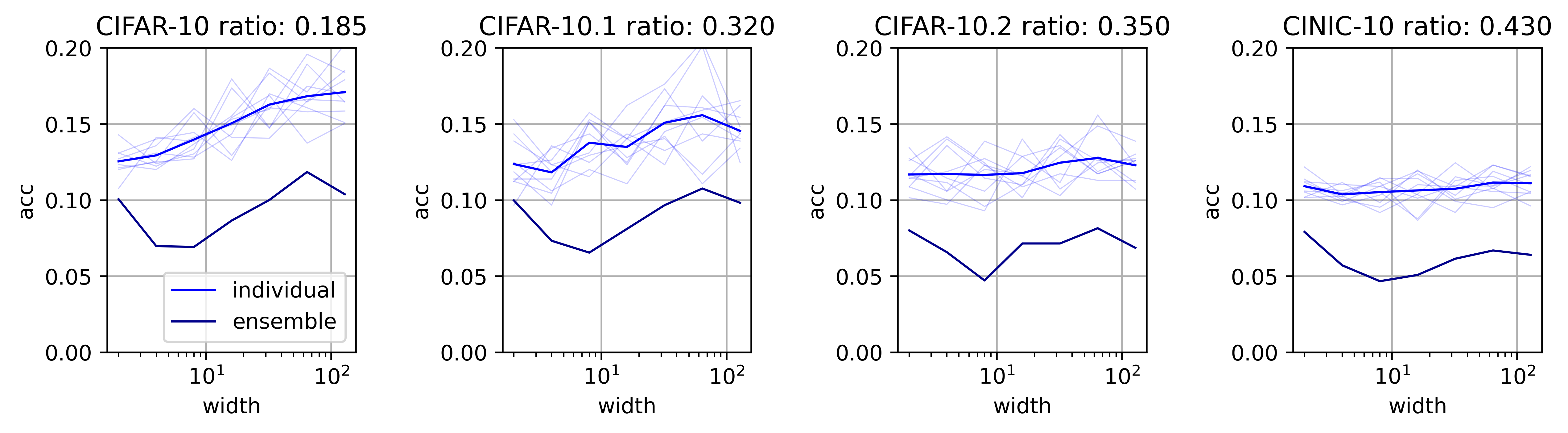} \\
    \vspace{-0.2cm}
    {\scriptsize \textbf{(c)} Hard/Ambigious subpopulation} \\
    \vspace{-0.2cm}
    \caption{\label{fig:subpops-OOD} \textbf{LeNet performances across subpopoulations of CIFAR-10 type ID and OOD datasets.} \textbf{First-row:} Typically the easy subpopulation makes up the biggest portion of the dataset (see the subpopulation ratios) therefore dominates the overall accuracy. \textbf{Second-row:} Underparameterized networks exhibit gradual improvement in medium subpopulation but overparameterized networks exhibit a drop in the accuracy. \textbf{Third-row:} In the hard subpopulation, the networks exhibit random guessing behavior (around $\%10$) and the networks perform pretty much the same independent of the network complexity.
    }
\end{figure}

As expected, ensembling improves the average accuracy on all test datasets (see Fig.~\ref{fig:subpops-OOD}). Note that although in the easy subpopulation we collected the samples that are most offen correctly classified, we observe that even the overparameterized networks do not achieve perfect accuracy in this subpopulation. However, ensembling pushes up accuracies to very close to $1.00$ not only for overparameterized networks, but also for the medium-sized original Lenet with $w=16$. For the medium subpopulation, we observe that ensembling brings a larger gain in accuracy, possibly due to a contamination of this subpopulation by weak spurious correlations. Suprisingly, we observe that ensembling hurts the accuracies in the hard subpopulation! This is due to the mechanism of ensembling: that it implements the the decision of the majority of decision makers. In the case of the hard subpopulation, the majority of the networks are incorrect, therefore the ensemble makes the wrong decision, cancelling out the correct decision given by the minority of networks.
\pagebreak

\begin{figure}[b!]
  \begin{center}
    \includegraphics[width=0.7\textwidth]{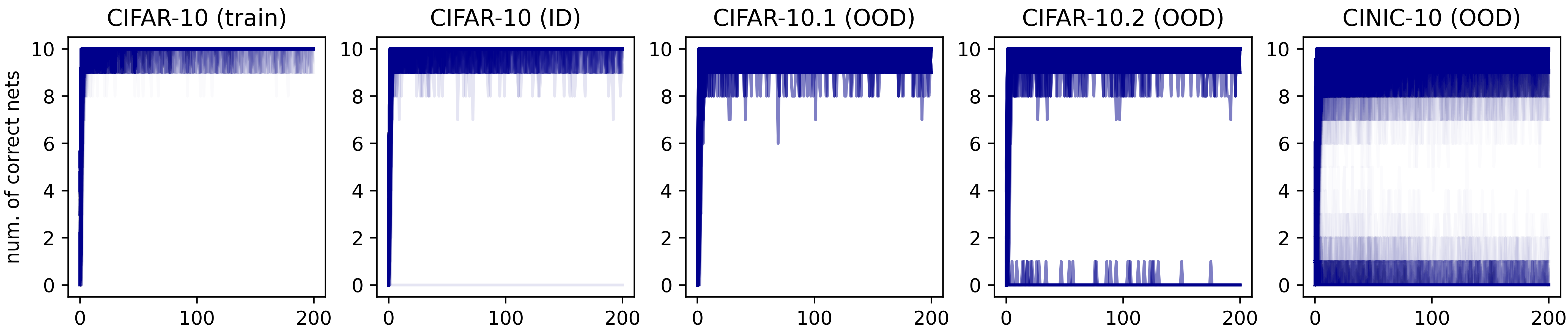}
    \includegraphics[width=0.7\textwidth]{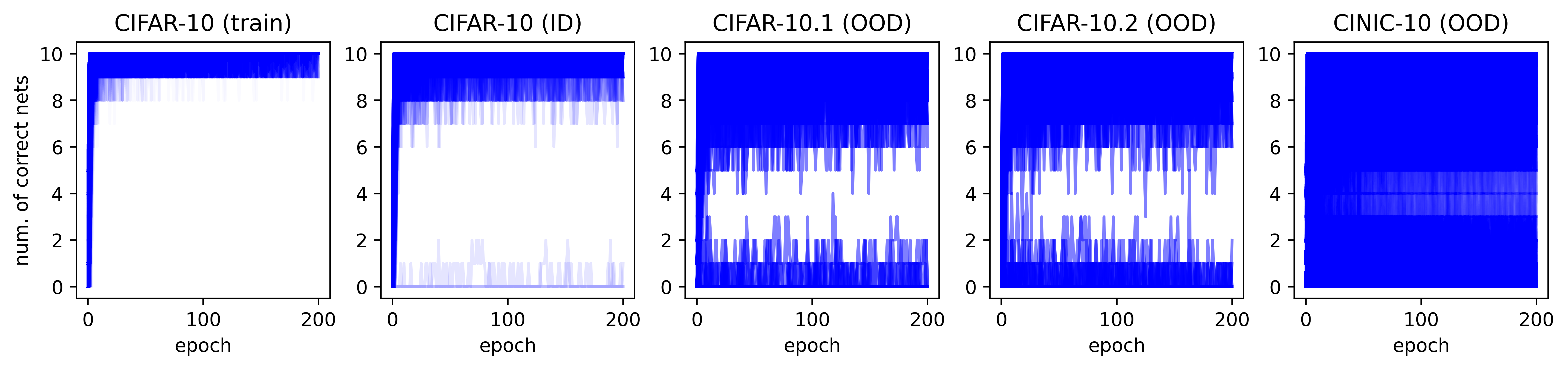}
  \end{center}
  \vspace{-0.3cm}
  \caption{\label{fig:dynamics-per-group-app} The easy samples in the low confusion group are rarely forgotten and hard samples are rarely classified correctly. Interestingly, for some easy images, not all networks get it right! First row represents first fifth of the samples in the lowest entropy group, the second row represents the second fifth. In the other groups we do not see a separation between easy and hard examples which suggests that the networks keep learning and forgetting these samples in a cyclic fashion (except for the training data where we observe all the samples are learnt/memorized similar to the first two rows. But there are accidental mistakes (in easy samples) and accidental learning (in hard samples) due to the randomness coming from the initialization and learning algorithm (the order of which the samples are learnt plays a significant role).}
 \end{figure}

\pagebreak

\section{Further images for visual inspection}

\begin{figure}[h!]
  \begin{center}
  \includegraphics[width=0.48\textwidth]{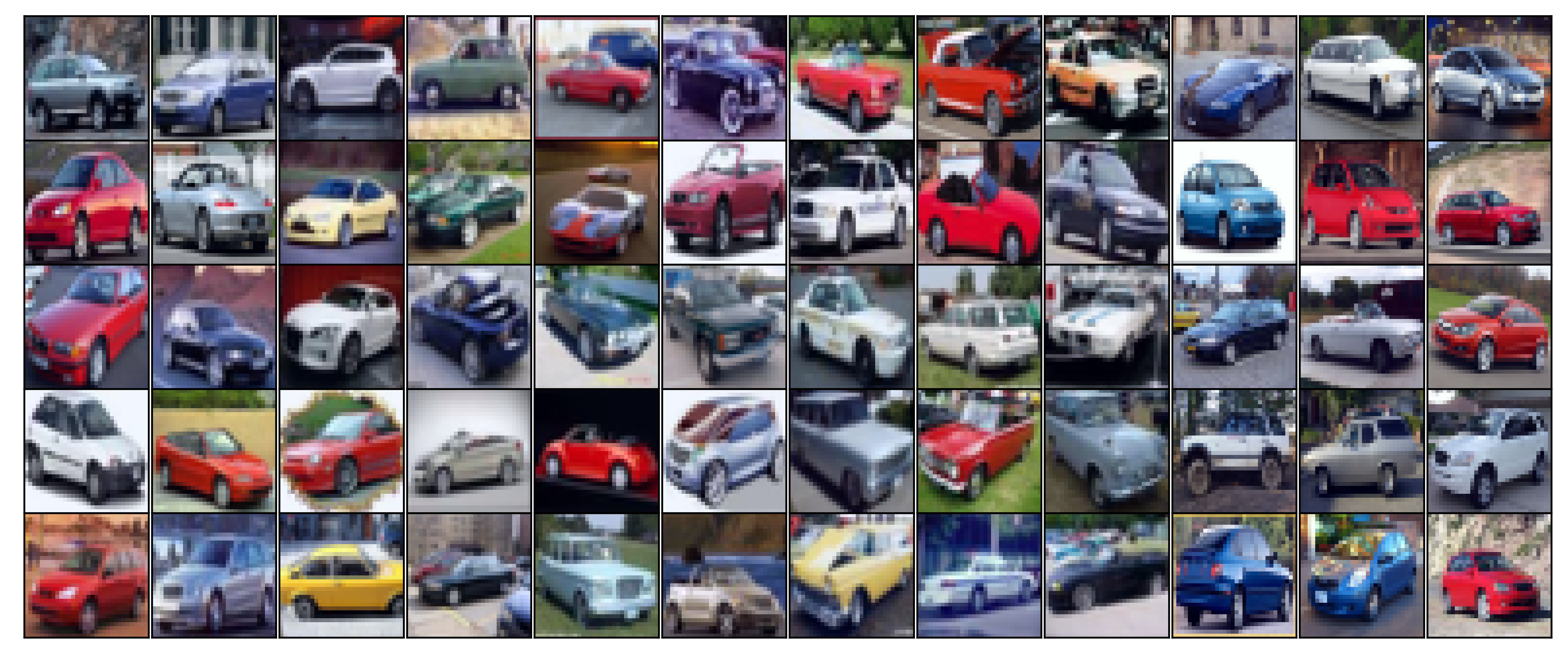} \hspace{3mm}
    \includegraphics[width=0.48\textwidth]{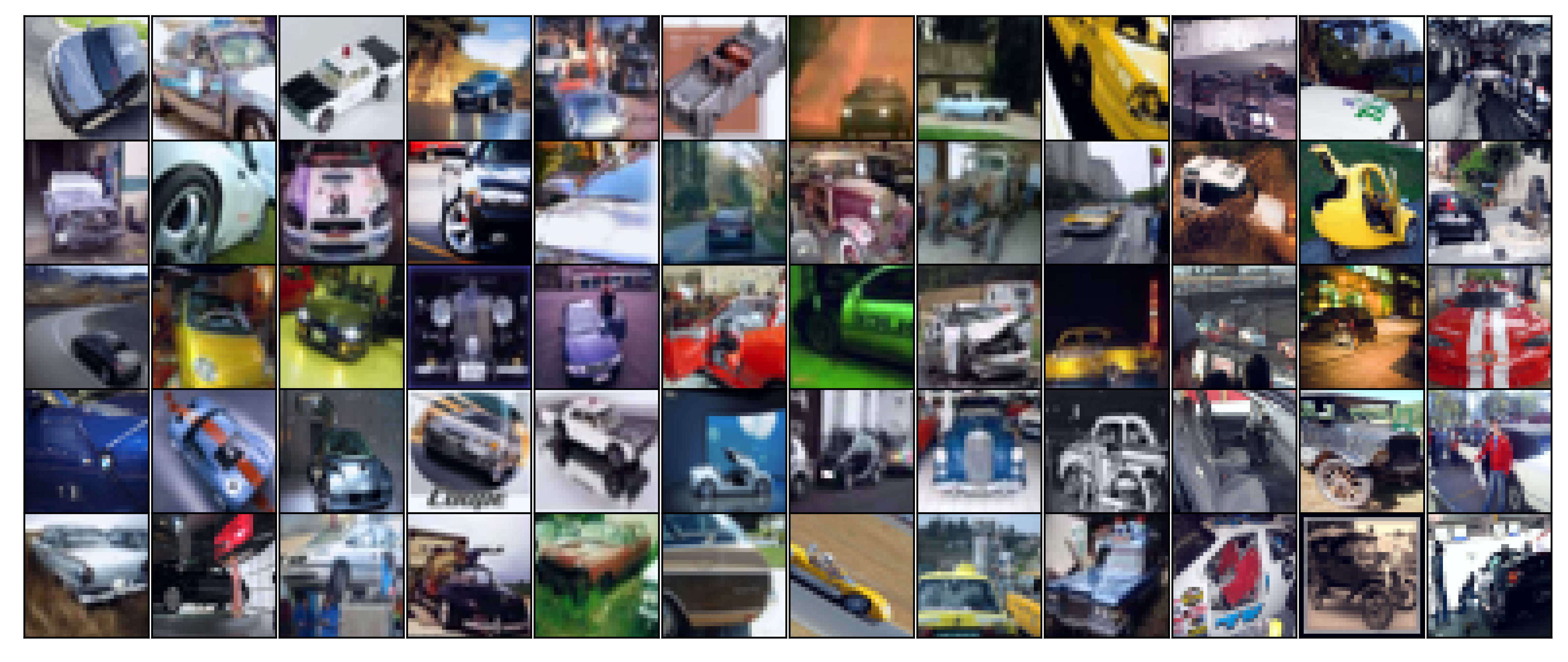}
    \includegraphics[width=0.48\textwidth]{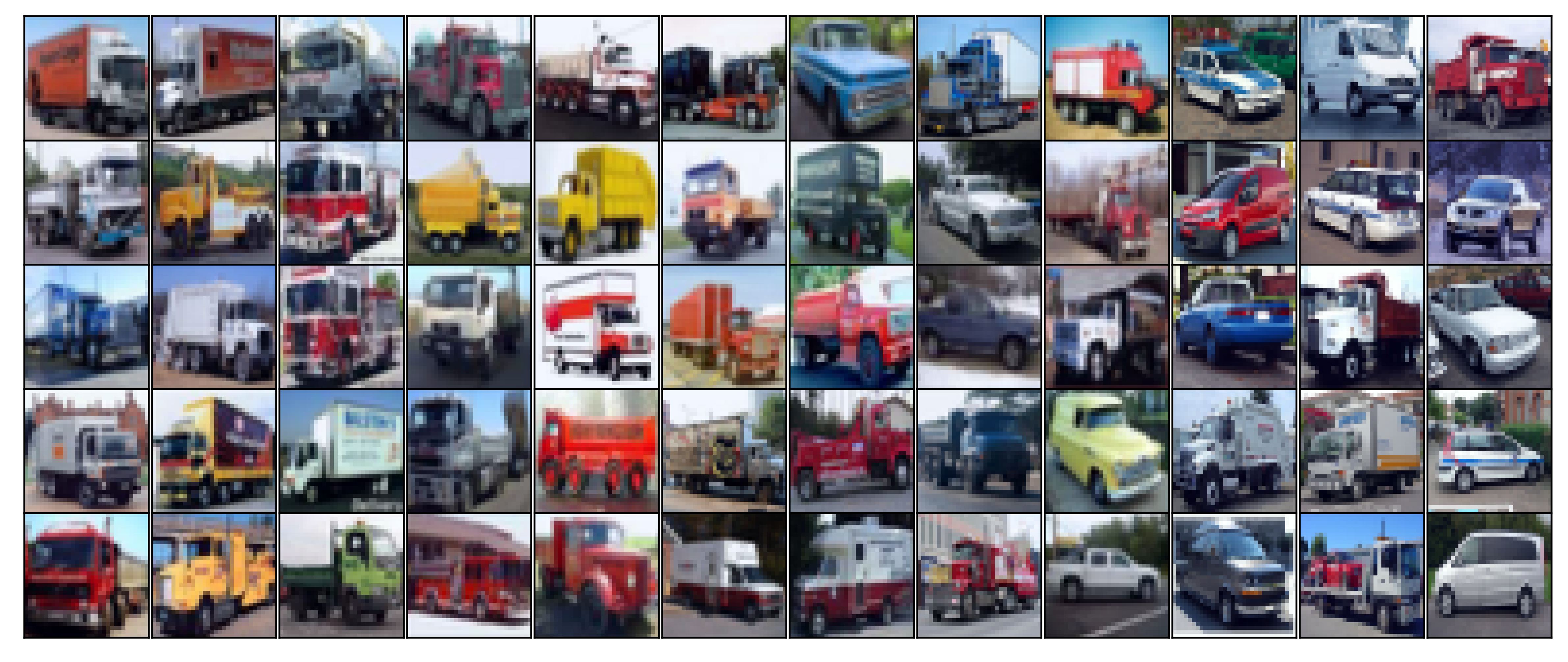} \hspace{3mm}
    \includegraphics[width=0.48\textwidth]{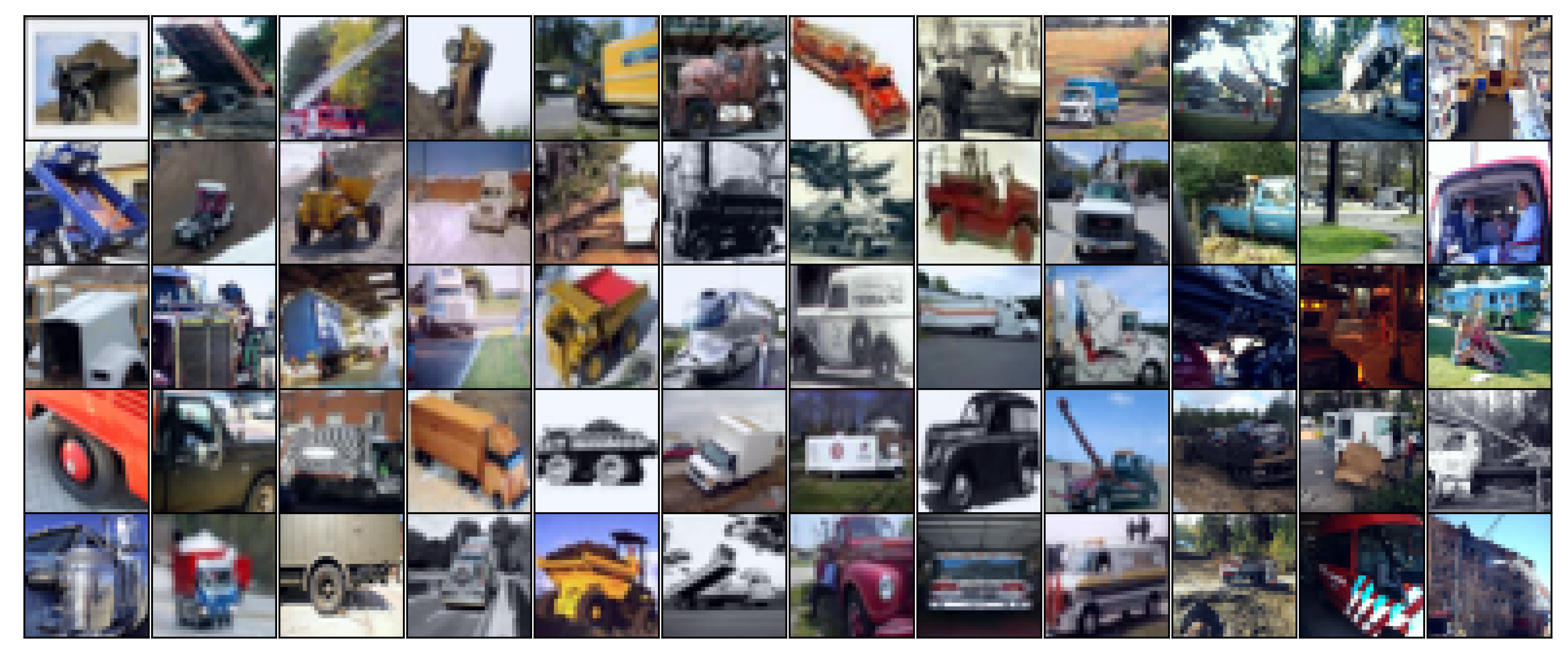}
      \includegraphics[width=0.48\textwidth]{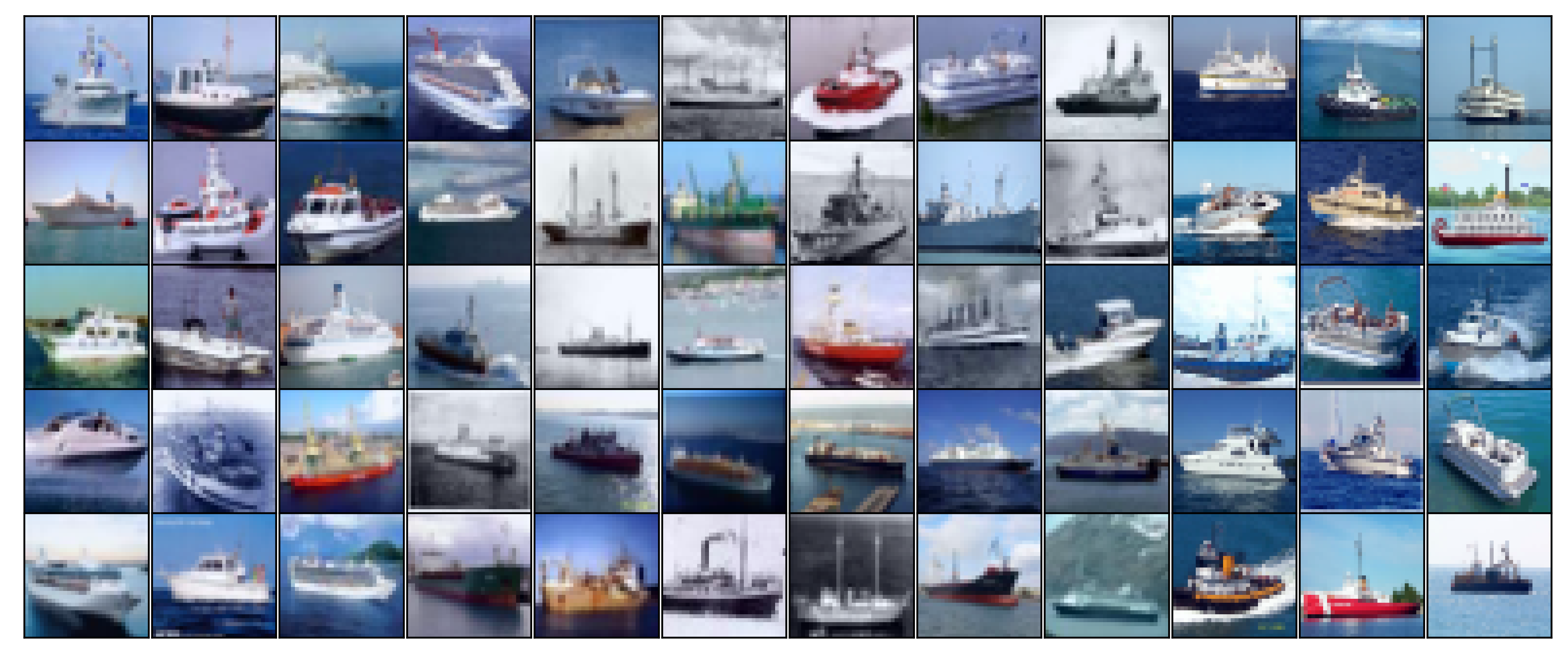} \hspace{3mm}
    \includegraphics[width=0.48\textwidth]{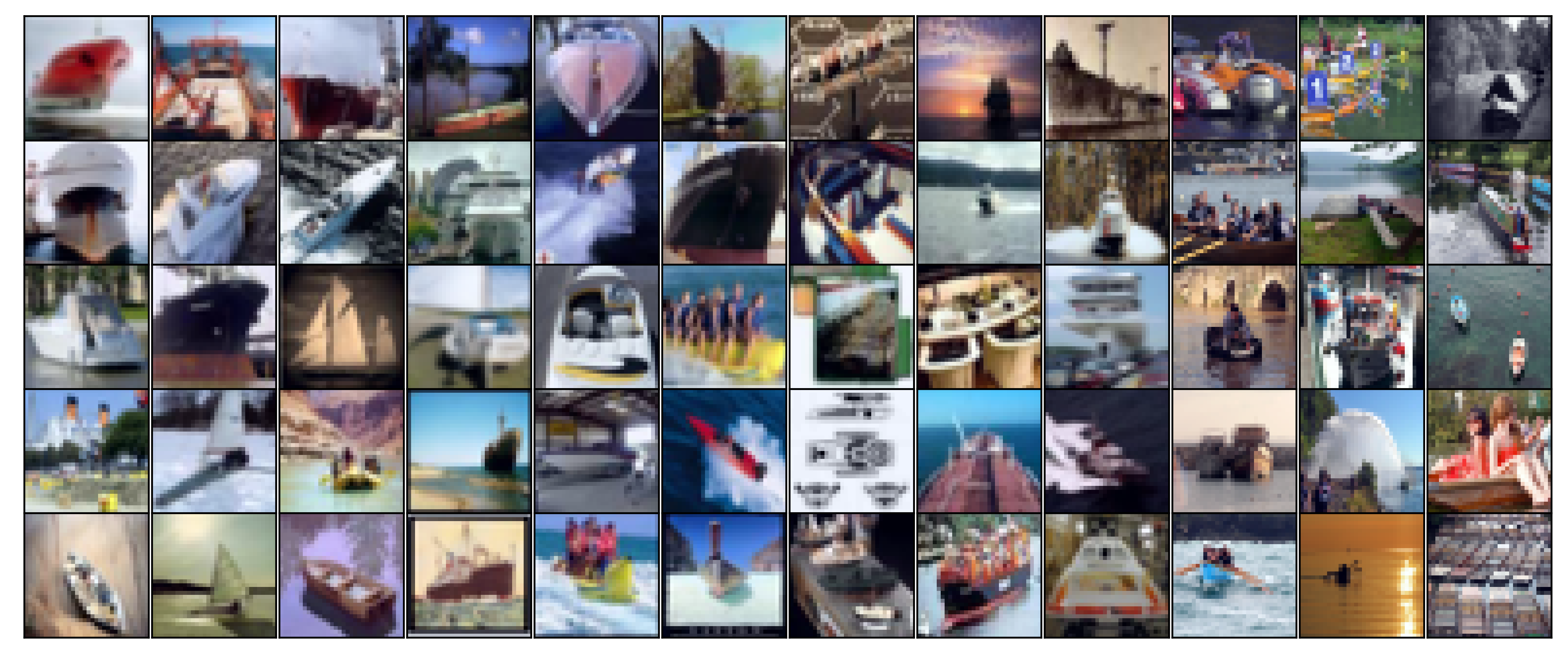}
  \includegraphics[width=0.48\textwidth]{figs/01pics-plane-low.png} \hspace{3mm}
    \includegraphics[width=0.48\textwidth]{figs/01pics-plane-high.png}
      \includegraphics[width=0.48\textwidth]{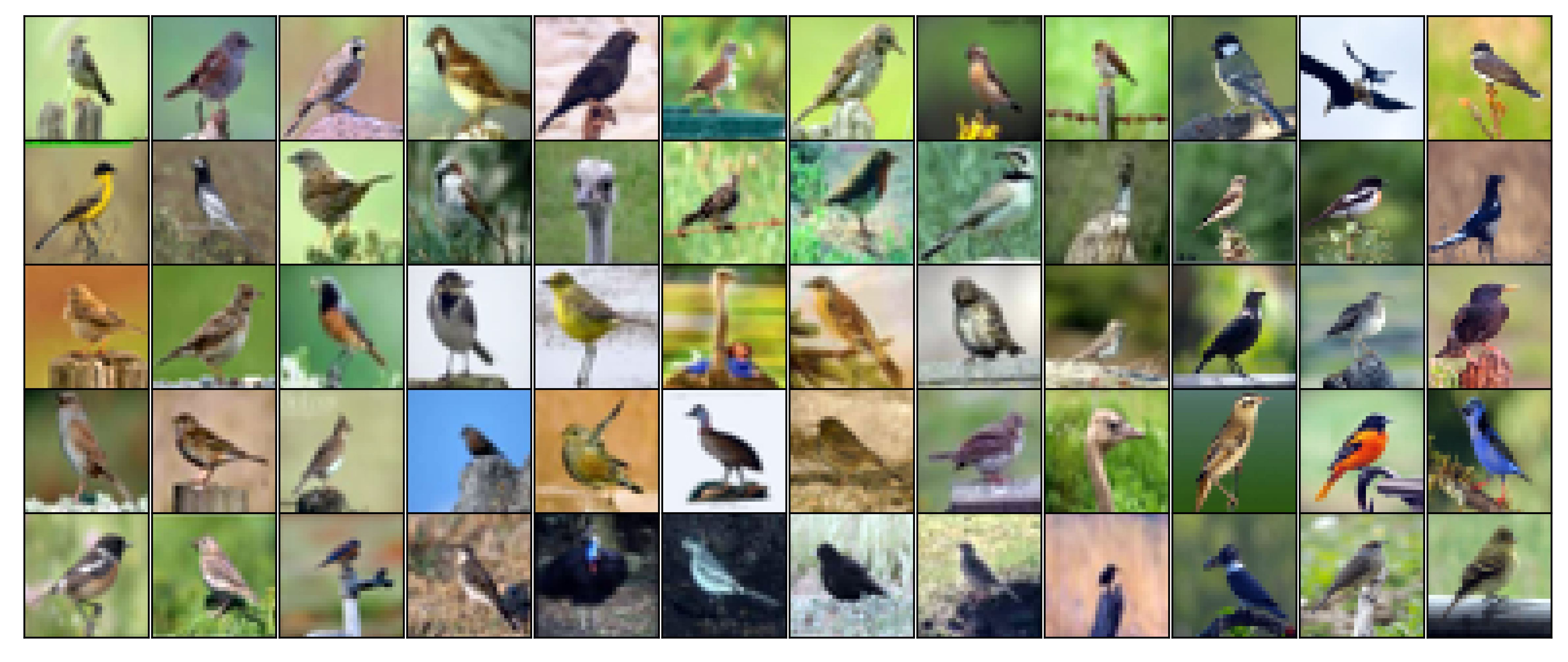} \hspace{3mm}
    \includegraphics[width=0.48\textwidth]{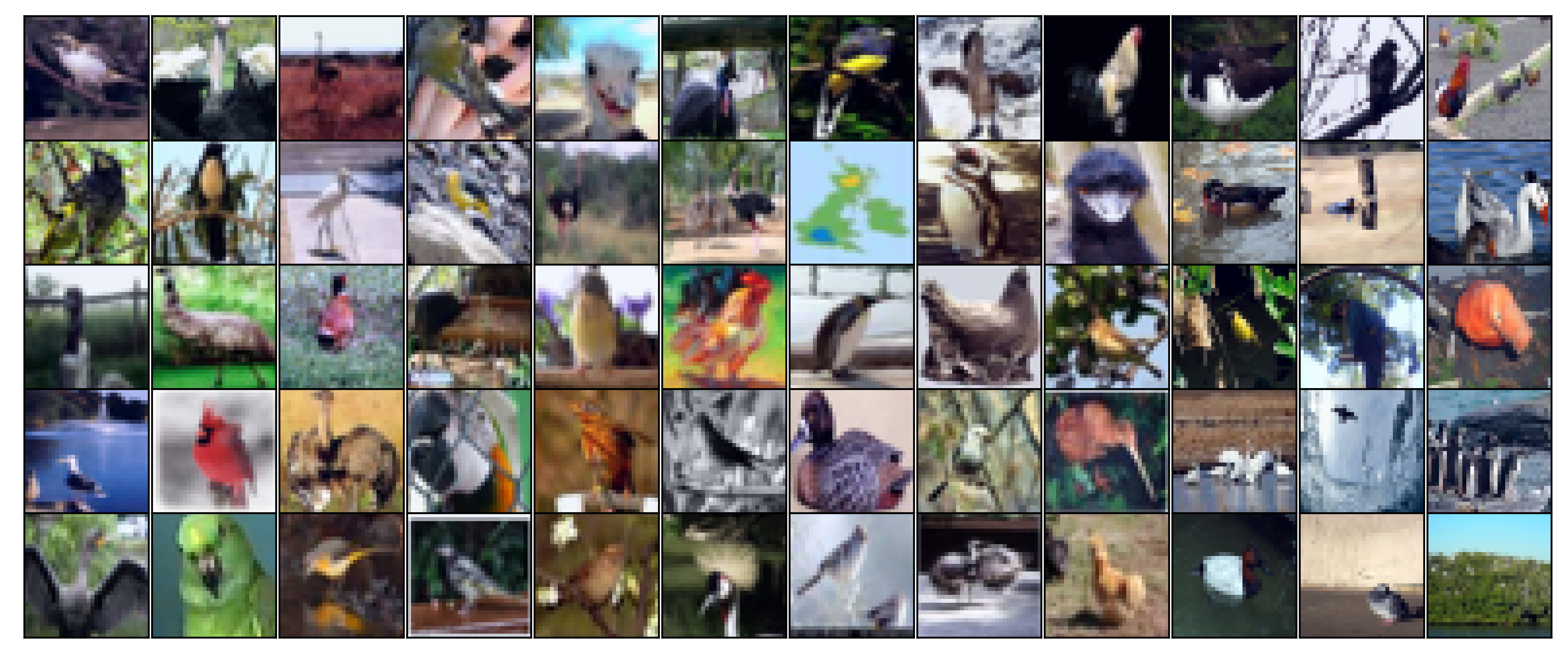}
  \end{center}
  \vspace{-0.3cm}
  \caption{\label{fig:low-high-entropy-app1} \textbf{Vehicle classes \& the bird class: Car, truck, ship, plane, bird.}
  Left: lowest confusion score images; Right: highest confusion score images from the CIFAR-10 test dataset (ID) and three OOD datasets (CIFAR-10.1, CIFAR-10.2, and CINIC-10); three columns for each test dataset.
  For the lowest confusion score images, we observe class-specific features such as wheels, front glass for the car class; high vehicle, truck trailer for the truck class; deck, main mast for the ship class; wings, tail for the plane class; beak, claws for the bird class. Also for the lowest confusion score images, we observe pronounced class-specific spurious correlations that match the labels of the classes, and limited amount of weak spurious correlations. For the highest confusion score samples, class-specific features and spurious correlations are either not fully pronounced and/or some features are missing, whereas weak spurious correlations such as gray-scale and white background can be present.
  }
\end{figure}

\begin{figure}[t!]
  \begin{center}
  \includegraphics[width=0.48\textwidth]{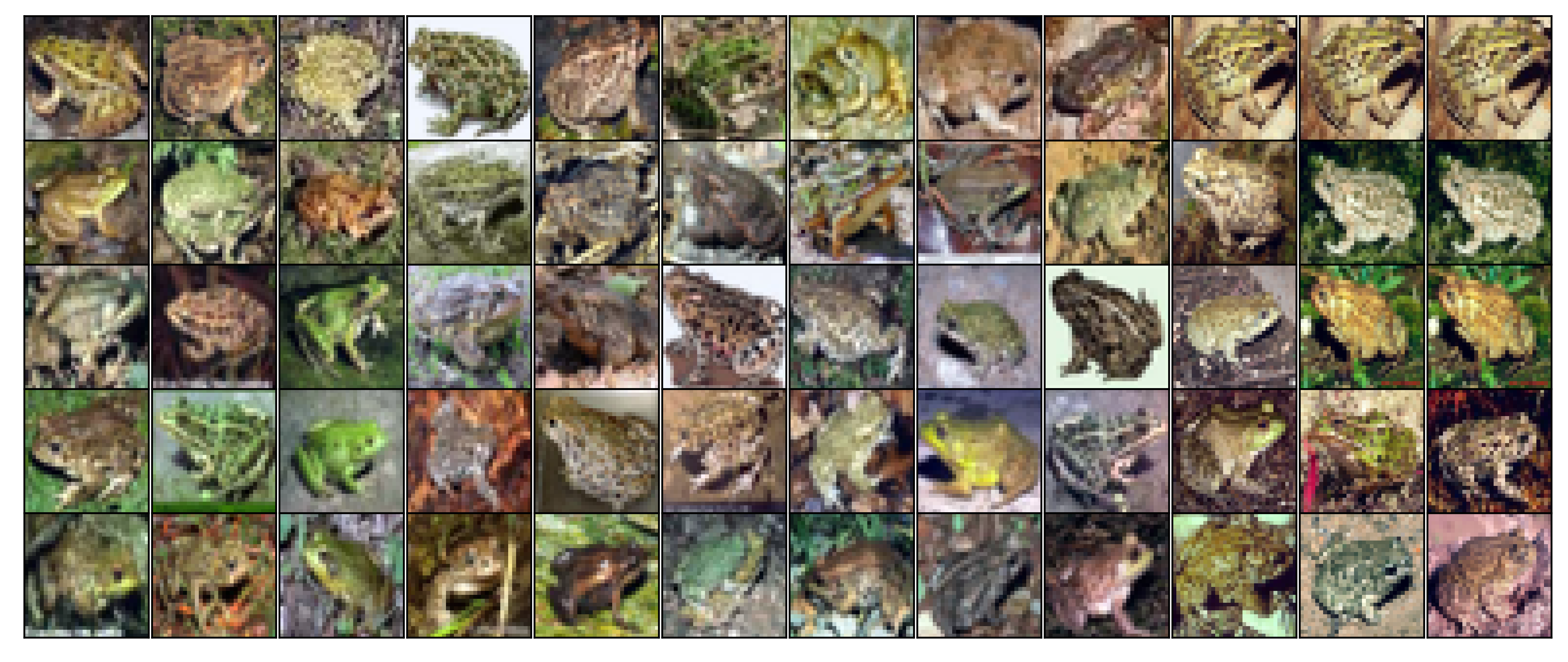} \hspace{3mm}
    \includegraphics[width=0.48\textwidth]{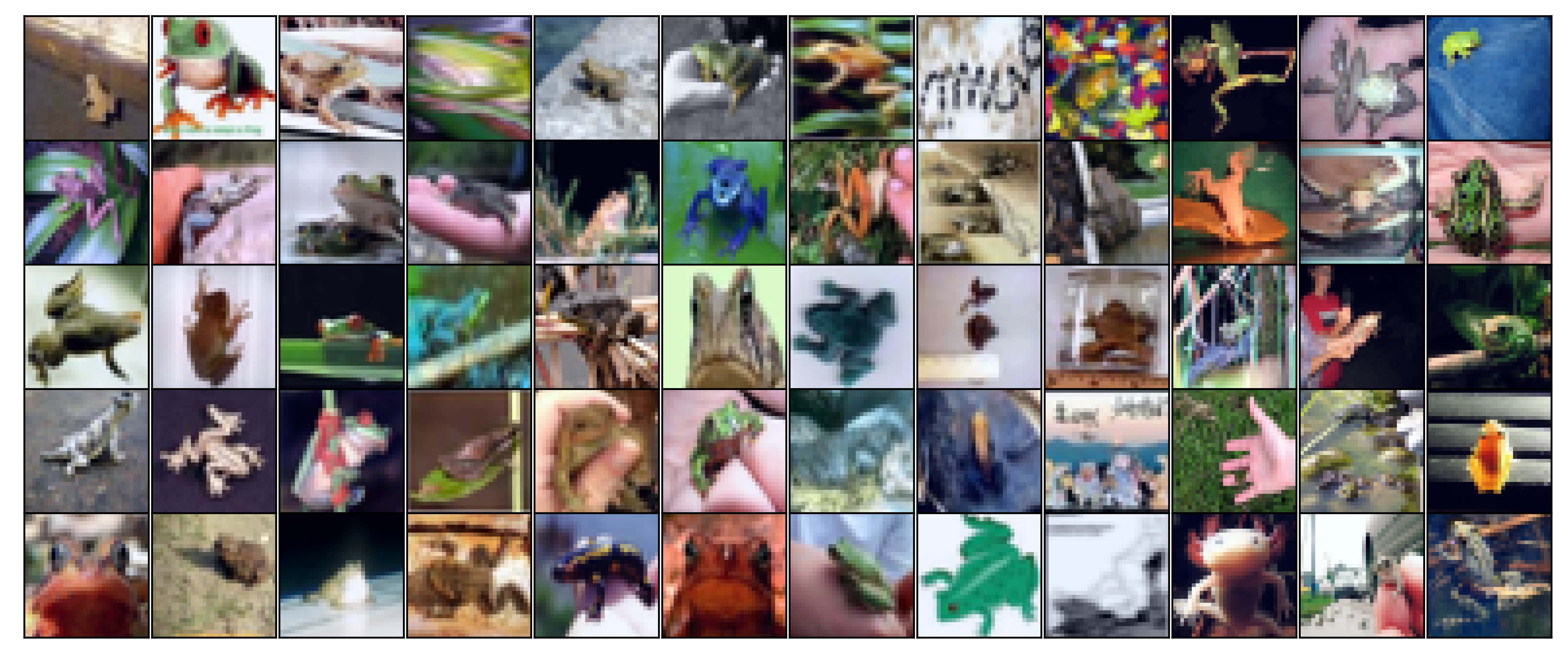}
    \includegraphics[width=0.48\textwidth]{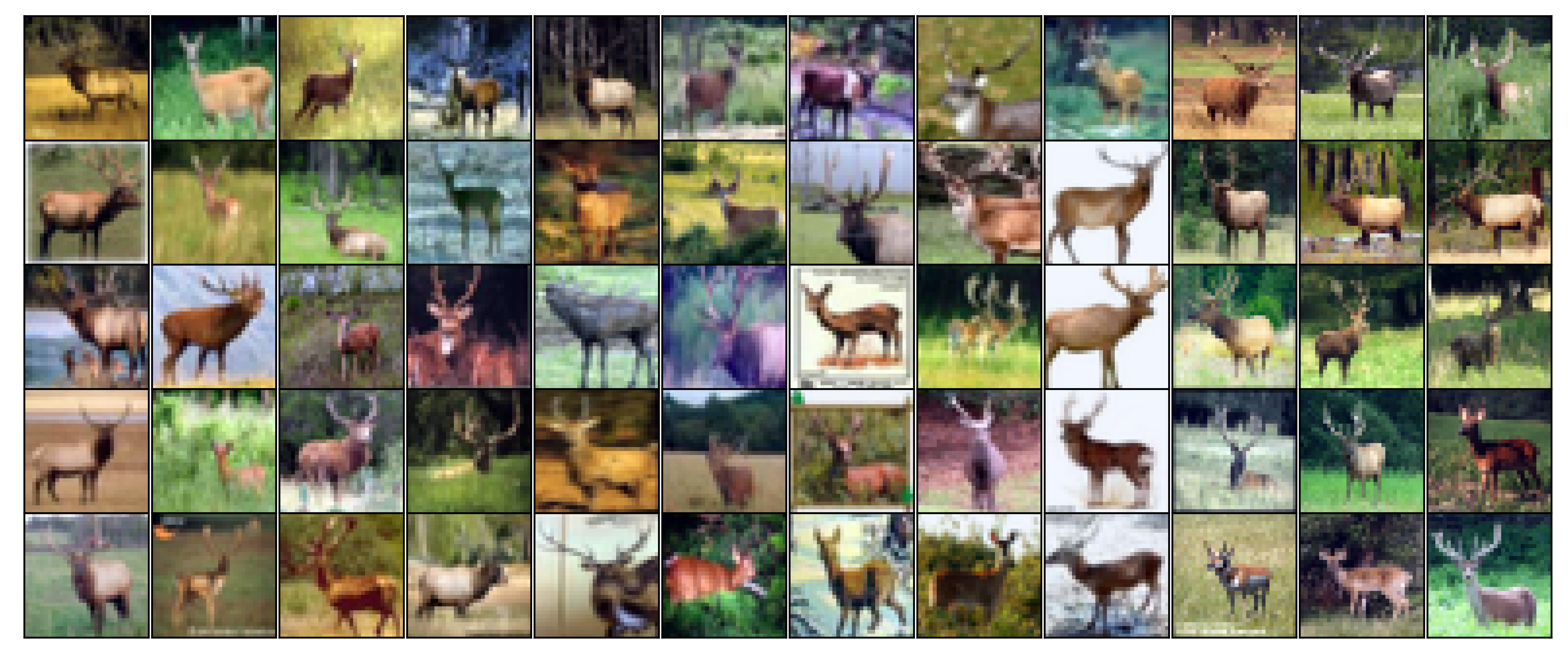} \hspace{3mm}
    \includegraphics[width=0.48\textwidth]{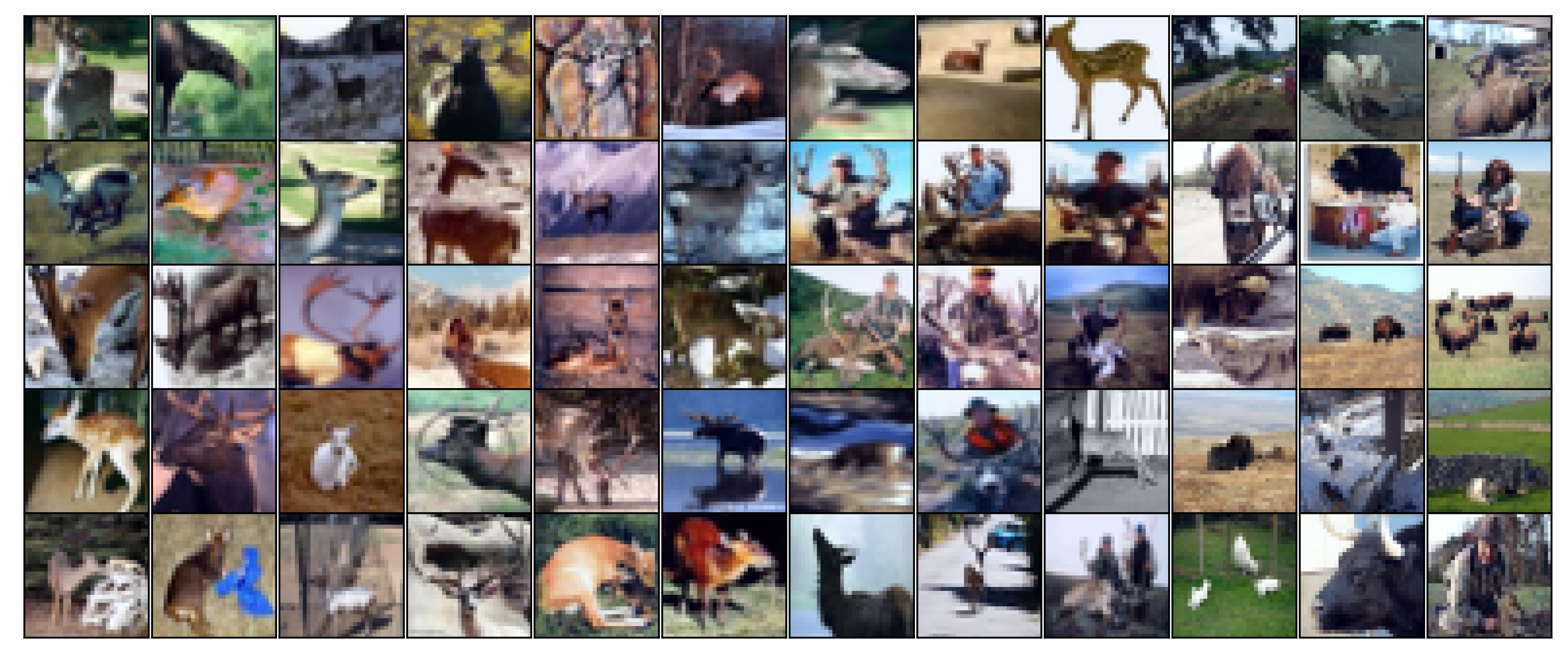}
    \includegraphics[width=0.48\textwidth]{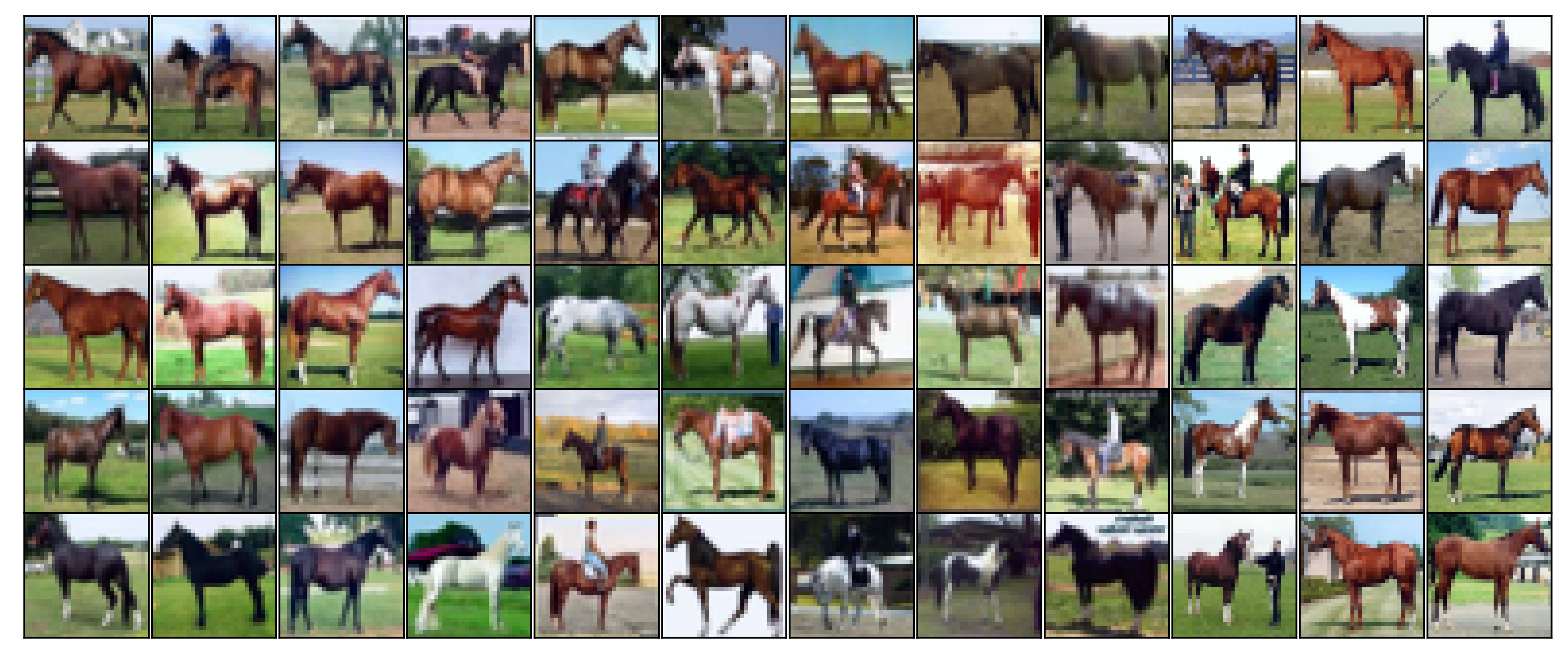} \hspace{3mm}
    \includegraphics[width=0.48\textwidth]{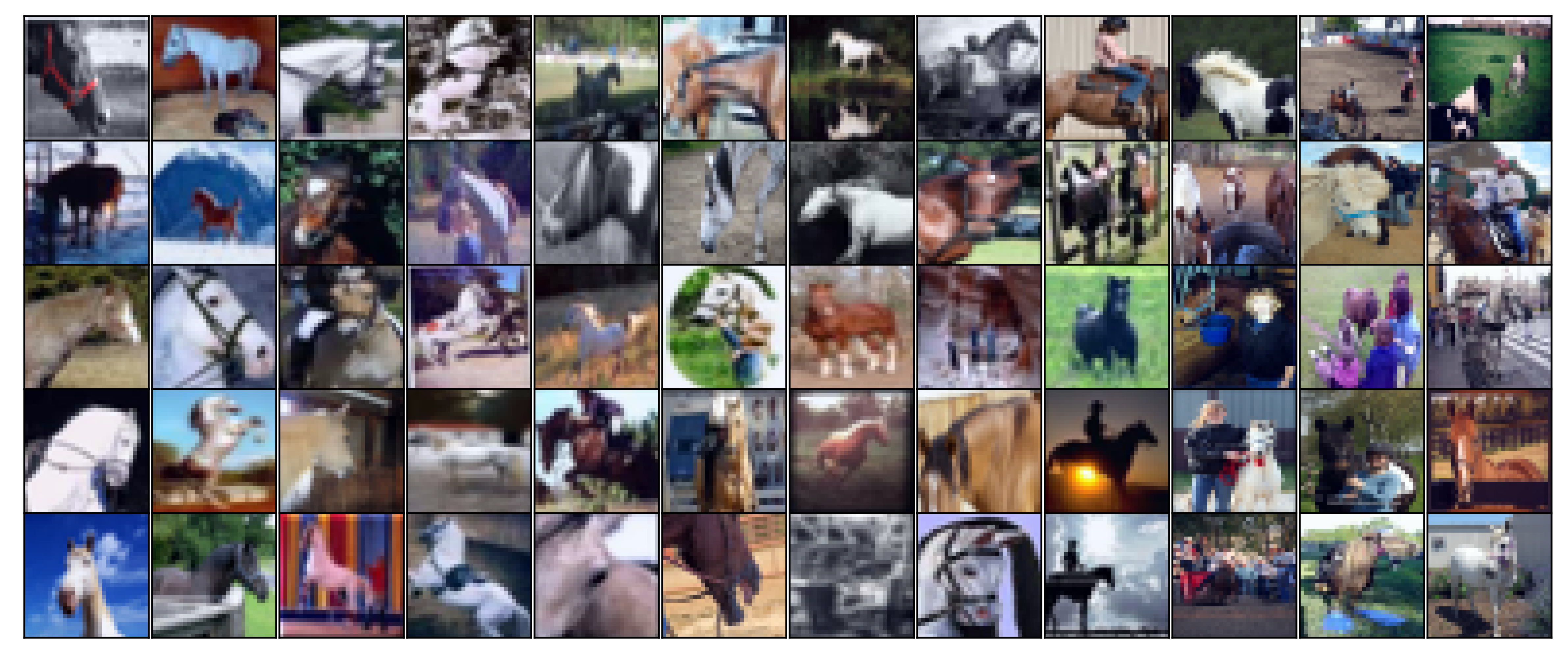}
    \includegraphics[width=0.48\textwidth]{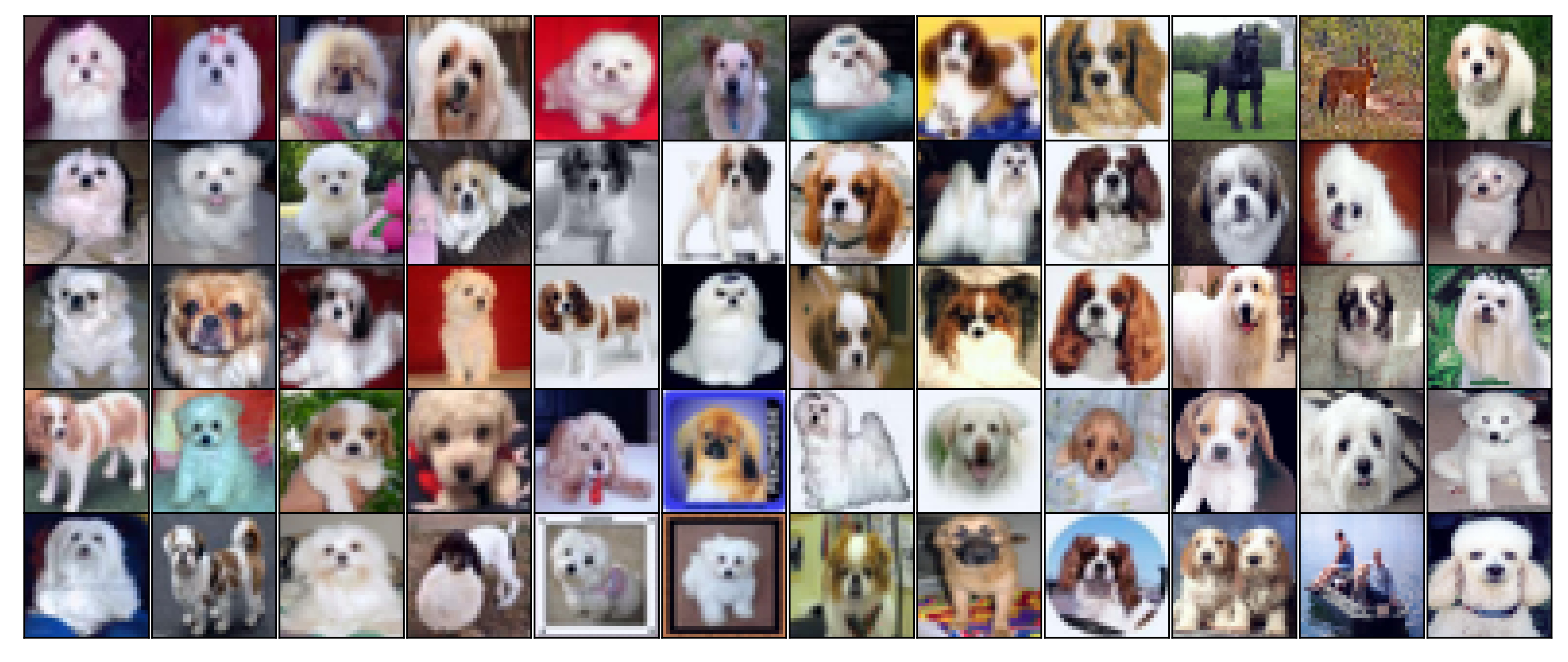} \hspace{3mm}
    \includegraphics[width=0.48\textwidth]{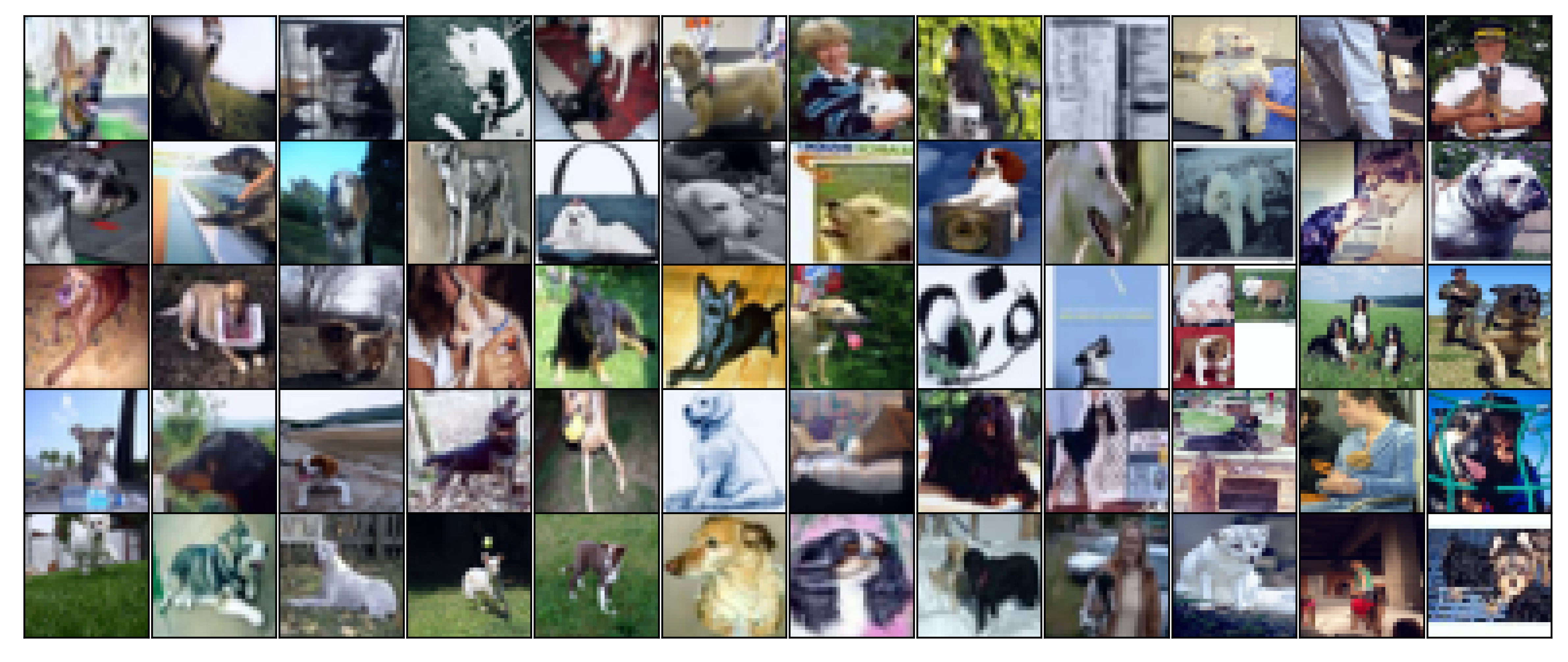}
    \includegraphics[width=0.48\textwidth]{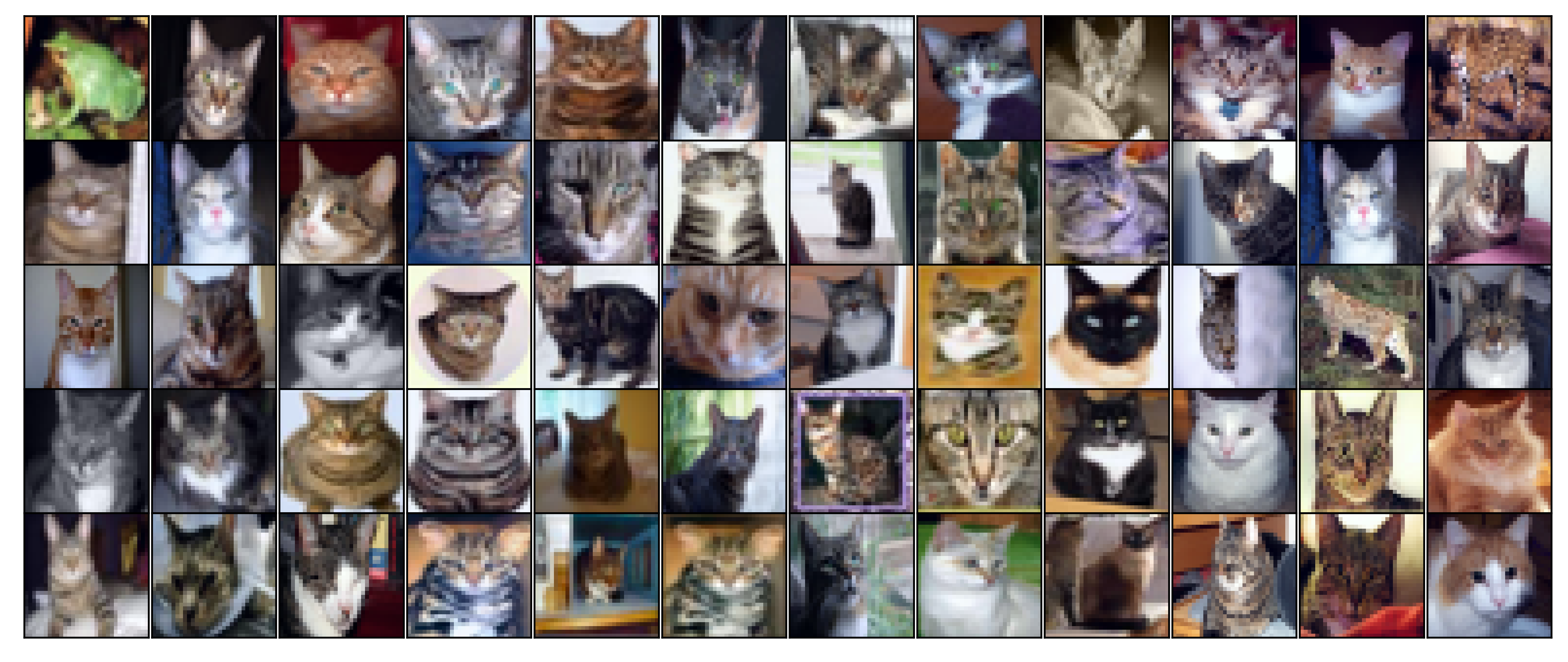} \hspace{3mm}
    \includegraphics[width=0.48\textwidth]{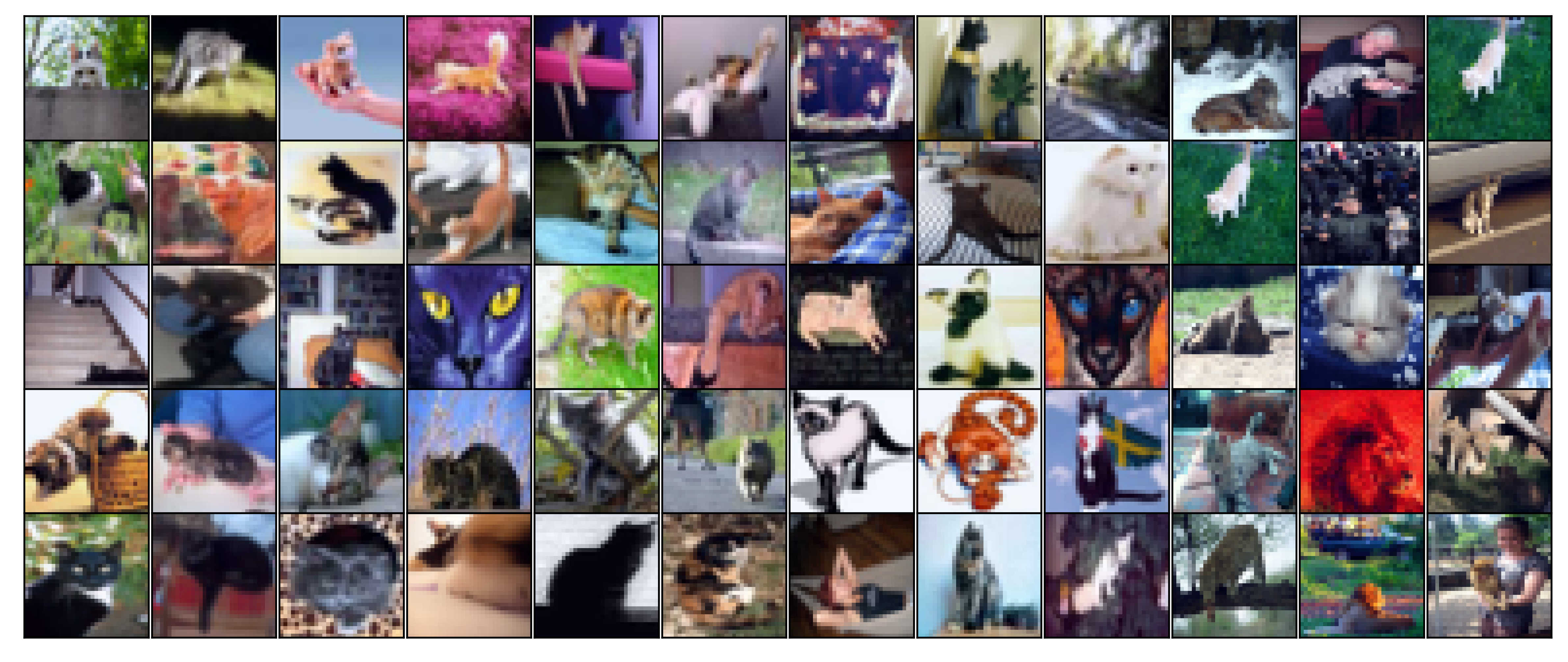}
  \end{center}
  \vspace{-0.3cm}
  \caption{\label{fig:low-high-entropy-app2} \textbf{Animal classes: Frog, deer, horse, dog, cat.}
  Left: lowest confusion score images; Right: highest confusion score images from the CIFAR-10 test dataset (ID) and three OOD datasets (CIFAR-10.1, CIFAR-10.2, and CINIC-10); three columns for each test dataset. For the lowest confusion score samples, we observe clear class-specific features and class-specific spurious correlations matching with the class labels. For the highest confusion score samples, some class-specific features or spurious correlations are either missing or not pronounced, and unusual color combinations for the given class may be present (i.e. weak spurious correlations).}
\end{figure}

\begin{figure}[t!]
  \begin{center}
    \includegraphics[width=0.24\textwidth]{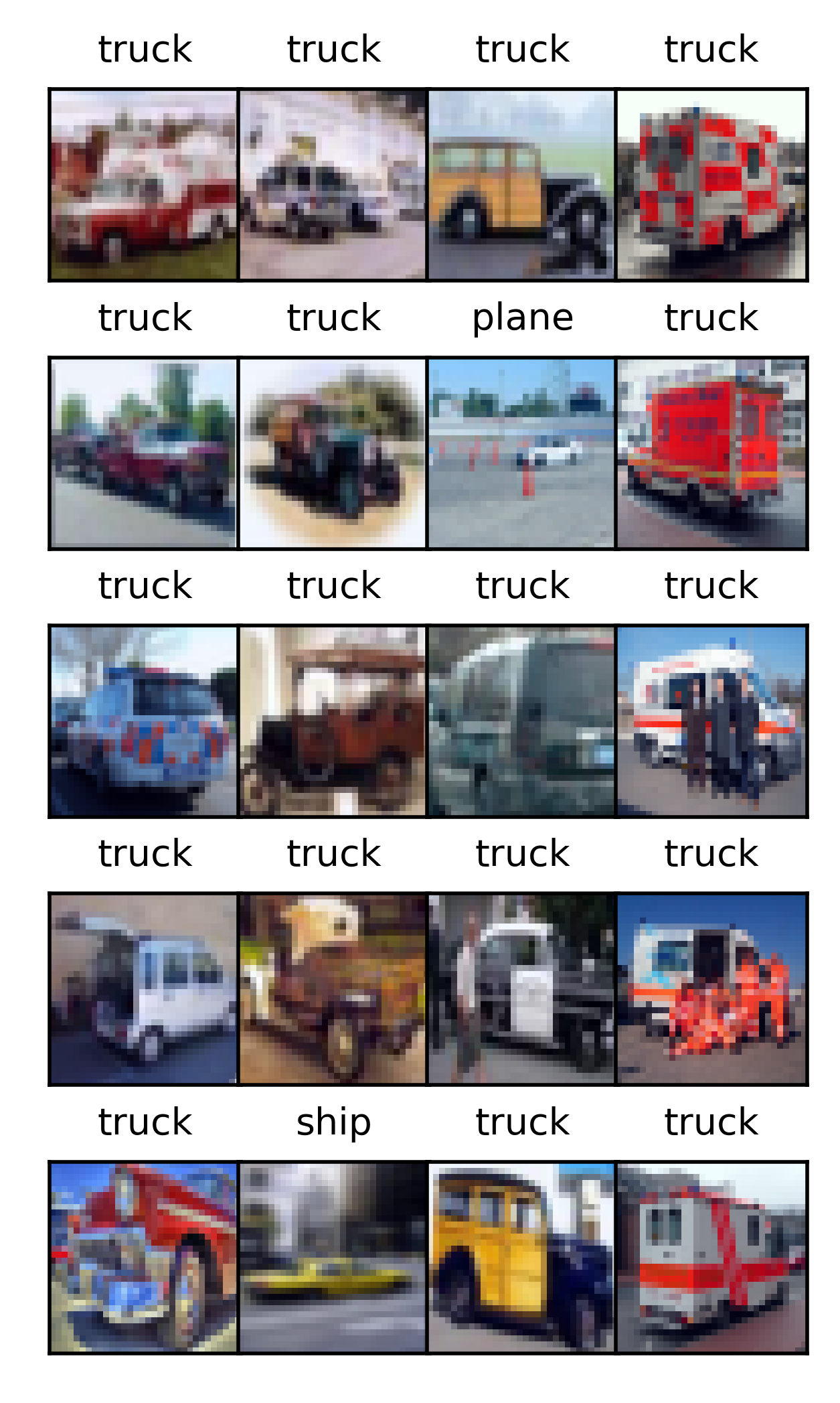}
    \includegraphics[width=0.24\textwidth]{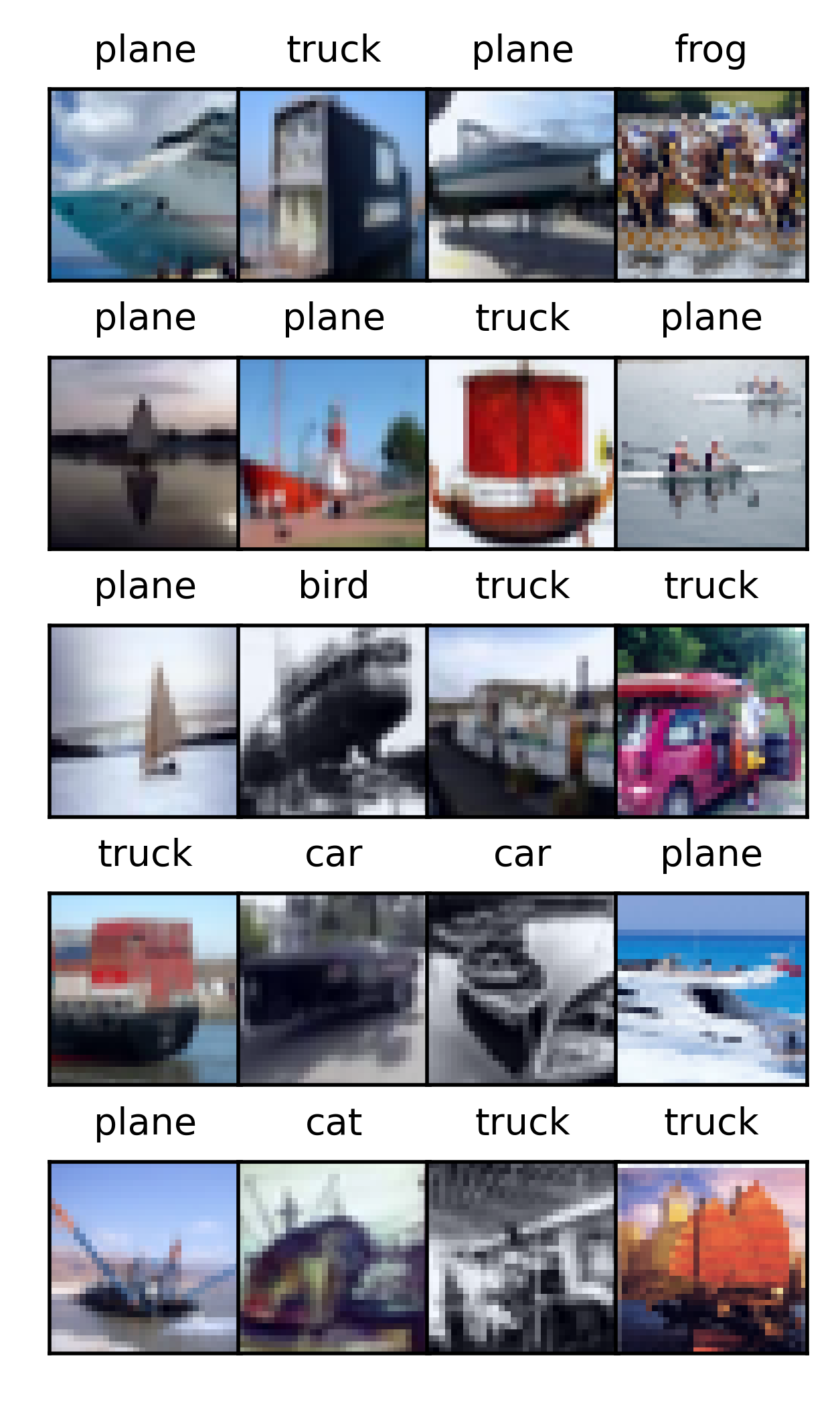}
    \includegraphics[width=0.24\textwidth]{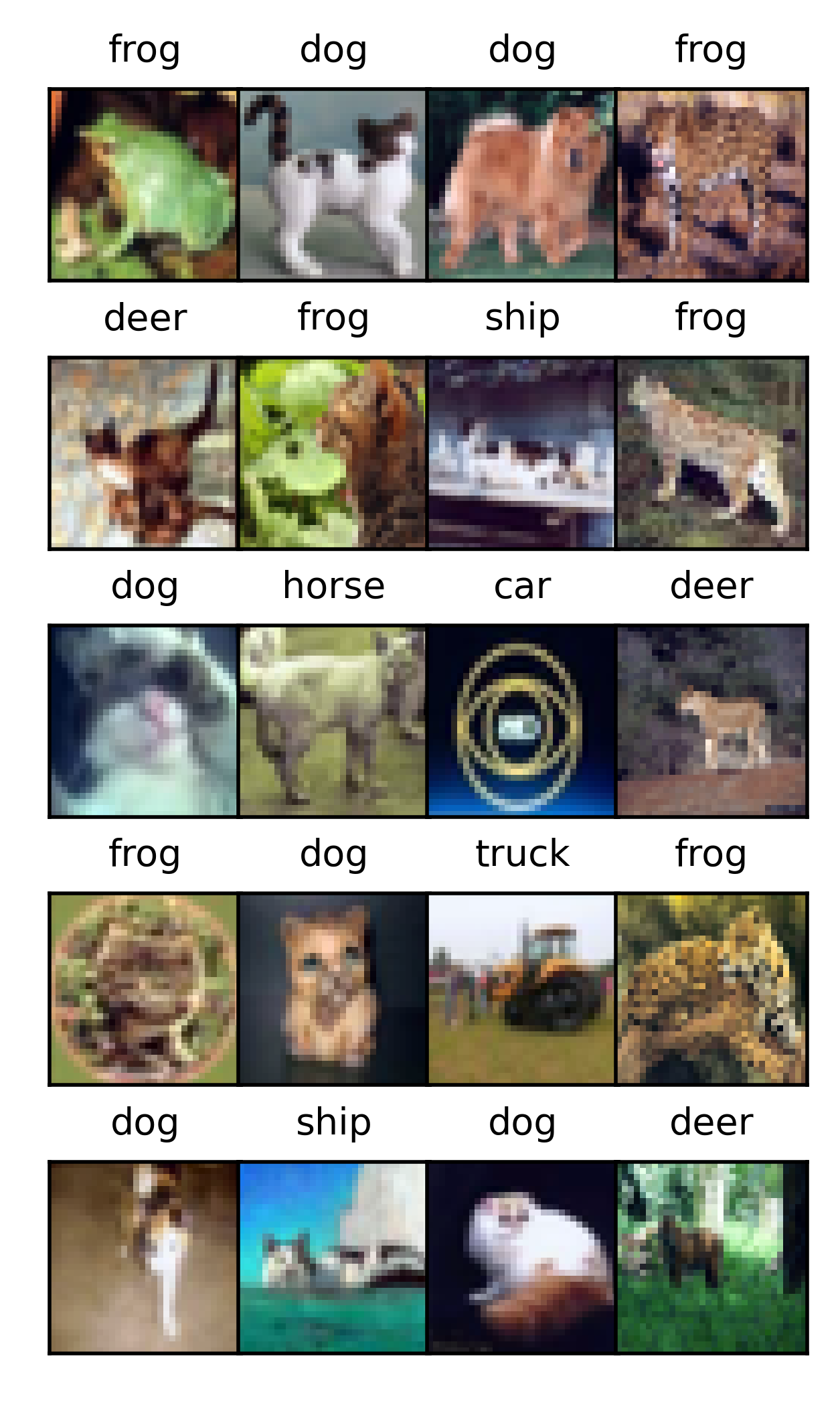} \\
    \textbf{{\scriptsize Car \hspace{3cm} Ship \hspace{3cm} Cat}} \\ \vspace{0.25cm}
    \includegraphics[width=0.24\textwidth]{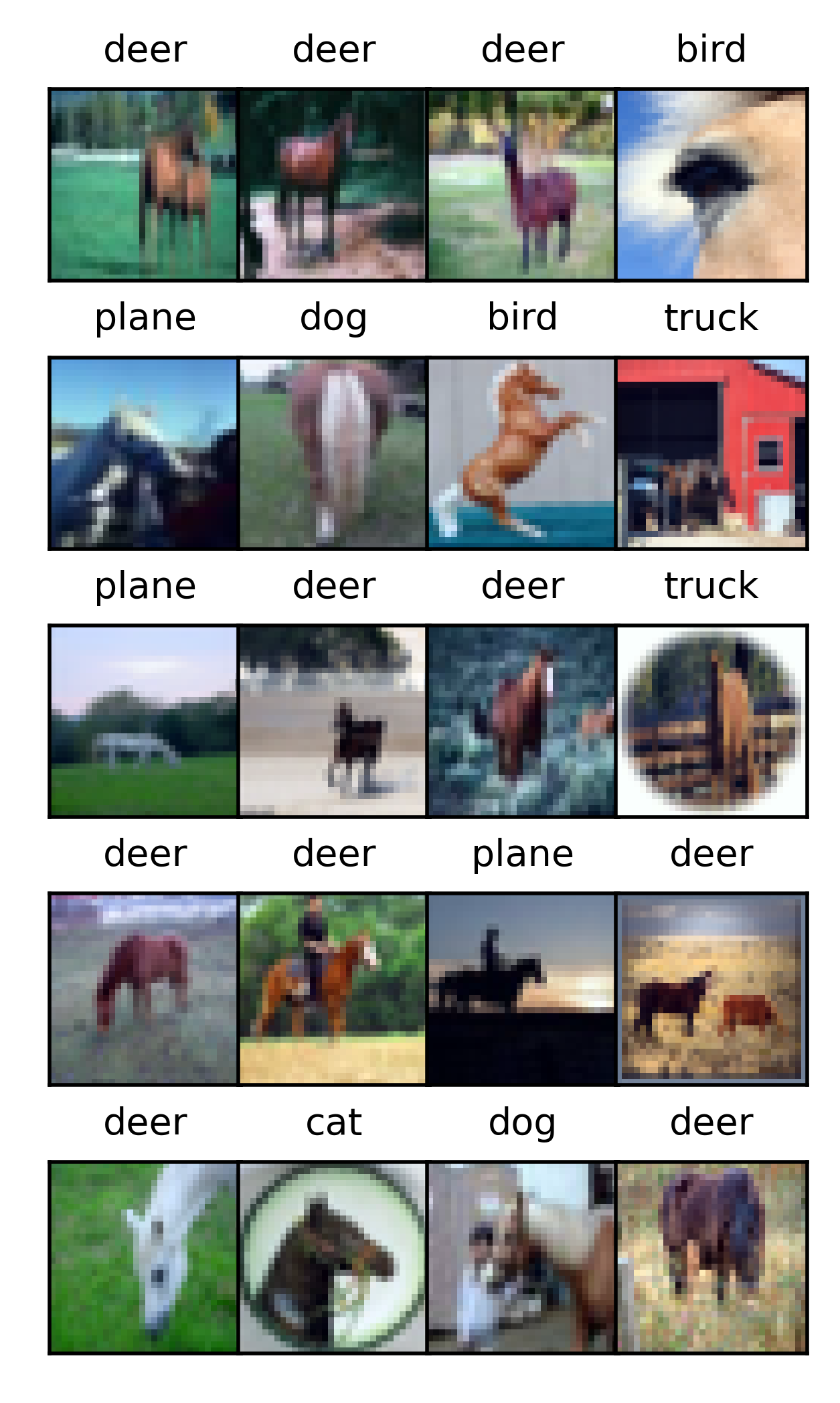}
    \includegraphics[width=0.24\textwidth]{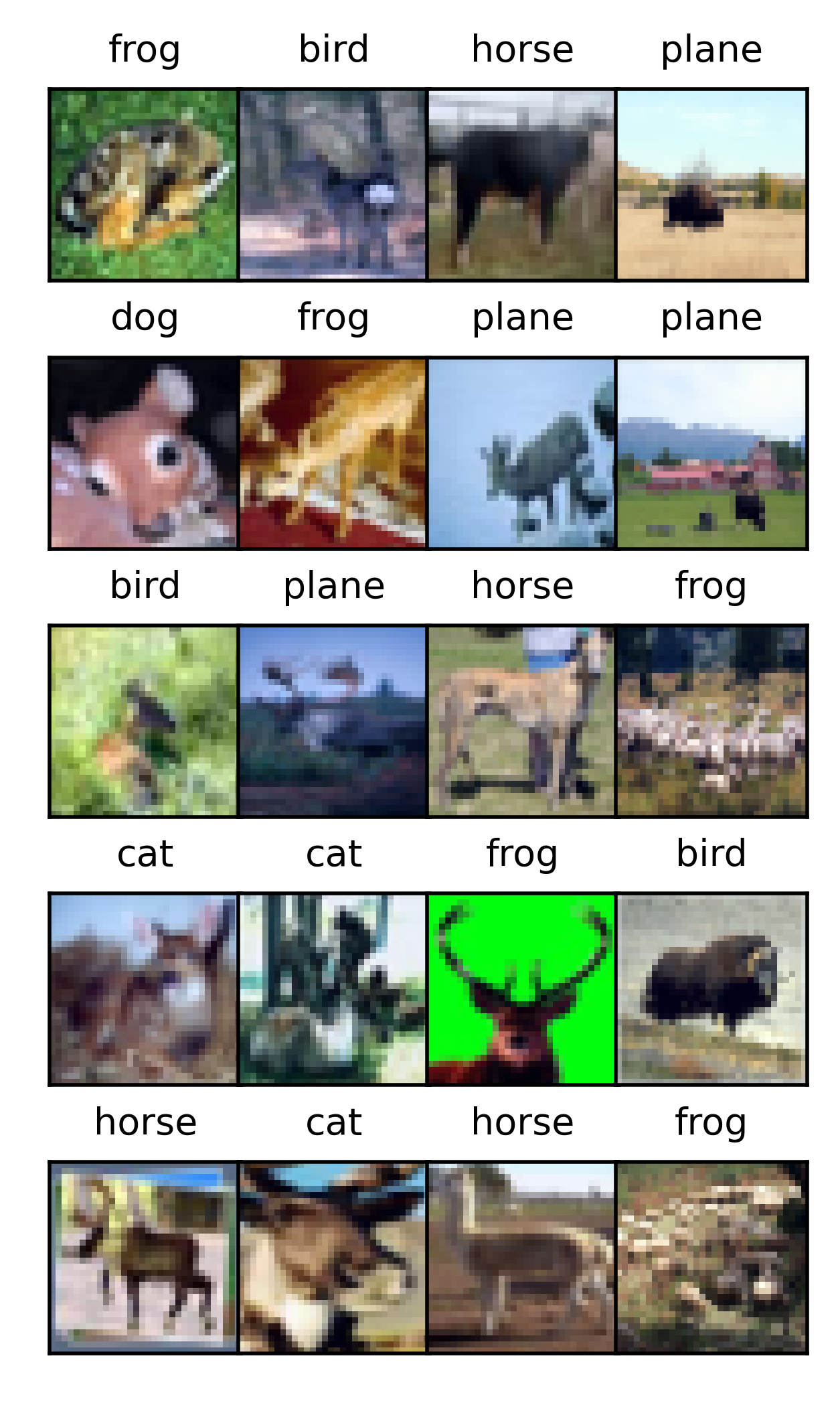}
    \includegraphics[width=0.24\textwidth]{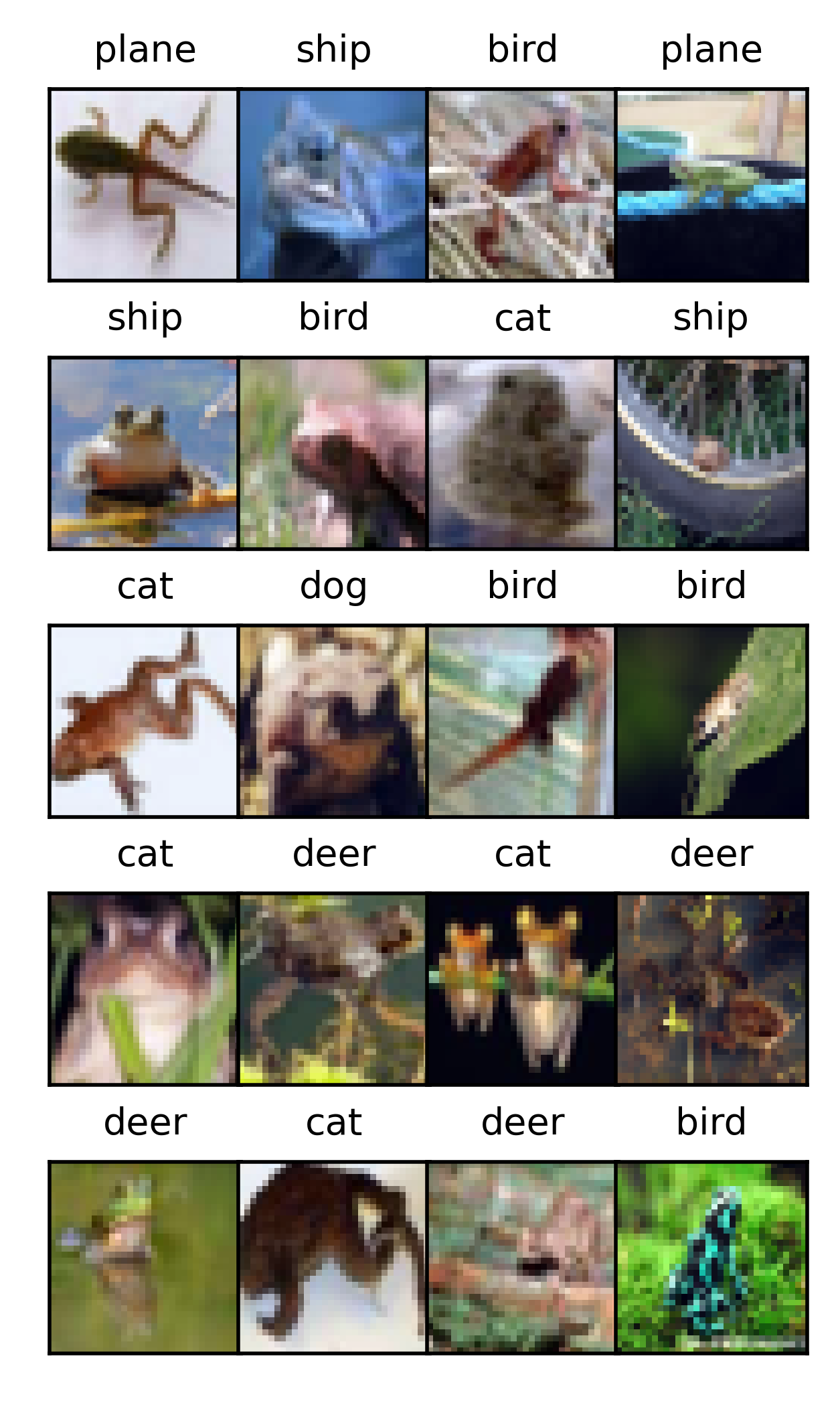} \\
    \textbf{{\scriptsize Bird \hspace{3cm} Deer \hspace{3cm} Frog}}
    \end{center}
  \vspace{-0.3cm}
  \caption{\label{fig:sp-corr-app} \textbf{Expansion of Fig.~\ref{fig:sp-corr} in the main.} Examples of class-specific spurious correlations (for the classes that are not presented in the main). The lowest confusion score samples that are misclassified by the majority of the ensemble are presented in this figure. Misclassifications in this group of images are mostly due to the presence of class-specific spurious correlations from the other classes than the label, and due to the lack of class-specific features. In the presence of clear class-specific features, the networks can overcome the effect of other class-specific spurious correlations present in the images and classify these correctly.}
\end{figure}

\end{document}